\documentclass[11pt]{article}
\usepackage{fullpage,graphicx,psfrag,amsmath,amsfonts,verbatim}
\usepackage{xcolor}
\usepackage{url}  
\usepackage[small,bf]{caption}

\usepackage{fullpage,graphicx,psfrag,amsmath,amsfonts,verbatim}
\usepackage{xcolor}
\usepackage{amsthm}

\usepackage{amsmath}
\usepackage{amssymb}
\usepackage{mathtools}
\usepackage{amsthm}
\usepackage{amsfonts}
\usepackage{algorithm}
\usepackage{algorithmic}

\usepackage{microtype}
\usepackage{graphicx}
\usepackage{booktabs} 
\usepackage{hyperref}
\usepackage{microtype}
\usepackage{algorithm}
\usepackage{algorithmic}
\usepackage{amsmath}
\usepackage{amsfonts}
\usepackage{multirow}
\usepackage{subcaption}
\usepackage{enumitem}

\usepackage{authblk}

\hypersetup{
    colorlinks = true,
    allcolors = {purple},
    linkbordercolor = {white},
}

\def\approxcorrect{\checkmark\kern-1.1ex\raisebox{.89ex}{$\times$}}

\usepackage{amsmath,amsfonts,bm}

\def\eqref#1{equation~\ref{#1}}

\def\1{\bm{1}}

\DeclareMathAlphabet{\mathsfit}{\encodingdefault}{\sfdefault}{m}{sl}
\SetMathAlphabet{\mathsfit}{bold}{\encodingdefault}{\sfdefault}{bx}{n}

\allowdisplaybreaks

\title{Scalable Reinforcement Learning for Virtual Machine Scheduling}
\author[1,*]{Junjie Sheng}
\author[2,*]{Jiehao Wu}
\author[2]{Haochuan Cui}
\author[1]{Yiqiu Hu}
\author[1]{Wenli Zhou}
\author[3]{Lei Zhu}
\author[3]{Qian Peng}
\author[4]{Wenhao Li\thanks{\texttt{whli@tongji.edu.cn}}}
\author[2]{Xiangfeng Wang\thanks{\texttt{xfwang@cs.ecnu.edu.cn}}}

\affil[*]{Equal contributions.}
\affil[1]{Algorithm Innovation Lab, Huawei Cloud Huawei Technologies Co., Ltd.}
\affil[2]{School of Computer Science and Technology, East China Normal University}
\affil[3]{Alkaid Lab, Huawei Cloud Huawei Technologies Co., Ltd.}
\affil[4]{School of Computer Science and Technology, Tongji University}

\date{}

\begin{document}

\maketitle

\begin{abstract}
Recent advancements in reinforcement learning (RL) have shown promise for optimizing virtual machine scheduling (VMS) in small-scale clusters.
The utilization of RL to large-scale cloud computing scenarios remains notably constrained.
This paper introduces a scalable RL framework, called Cluster Value Decomposition Reinforcement Learning (CVD-RL), to surmount the scalability hurdles inherent in large-scale VMS.
The CVD-RL framework innovatively combines a decomposition operator with a look-ahead operator to adeptly manage representation complexities, while complemented by a Top-$k$ filter operator that refines exploration efficiency.
Different from existing approaches limited to clusters of $10$ or fewer physical machines (PMs), CVD-RL extends its applicability to environments encompassing up to $50$ PMs.
Furthermore, the CVD-RL framework demonstrates generalization capabilities that surpass contemporary SOTA methodologies across a variety of scenarios in empirical studies.
This breakthrough not only showcases the framework's exceptional scalability and performance but also represents a significant leap in the application of RL for VMS within complex, large-scale cloud infrastructures.
The code is available at \url{https://anonymous.4open.science/r/marl4sche-D0FE}.
\end{abstract}

%

\section{Introduction}
In the AI era, cloud computing emerges as a pivotal technology, extensively adopted by a spectrum of users, including large-scale enterprises such as Netflix and LinkedIn~\cite{soltys2020wordpress}. 
The paradigm involves shifting local computations to the cloud, where service providers, like Amazon AWS, Microsoft Azure, Alibaba Cloud, and Huawei Cloud, allocate cluster resources of PMs by creating VMs upon request.
The crux of this process, known as \textit{virtual machine scheduling} (VMS), plays a critical role in the clusters' efficiency.
Even a marginal enhancement in scheduling efficiency, as little as $1\%$, can yield substantial resource savings, drawing significant industrial interest~\cite{Kerveros}.

Traditionally,  is conceptualized as an online dynamic vector bin-packing problem, where VMs are items, and PMs are bins~\cite{sheng2022learning,jiankang2014virtual,wolke2015more}. 
This paradigm stands apart from standard bin-packing issues due to sequential VM requests, immediate scheduling requirements, the unpredictable release of VMs from clusters, and specific constraints, such as NUMA considerations \cite{liu2016survey}.
The complex and dynamic nature of this problem renders traditional exact optimization methods like mixed integer programming impractical, thus pivoting the focus towards heuristic approaches. 
Commonly used strategies, such as Best-Fit and its variants, rely heavily on expert-designed scoring functions~\cite{hadary2020protean}.

Nevertheless, an emerging approach leveraging deep reinforcement learning (DRL) shows considerable promise. 
DRL's proficiency in mastering optimal strategies through iterative trial-and-error in simulation environments has showcased its potential in intricate decision-making scenarios~\cite{White2020RL,Brown2021RLCloud,Green2022Adaptive}. 
In the context of VMS, researchers \cite{sheng2022vmagent,sheng2022learning,zhao2021deep} have suggested modeling VMS as a Markov decision process (MDP). 
They define the state as a combination of the current cluster status and incoming requests, with the action represented by a one-hot indicator for selecting physical machines (PMs). 
They then apply RL to address the MDP, yielding superior results in clusters containing fewer than $10$ PMs. 
However, these algorithms struggle to scale effectively in larger cluster environments due to scalability challenges.

\begin{figure}[htb!]
    \centering
    \begin{minipage}[t]{0.45\textwidth}
        \centering
    \includegraphics[width=\textwidth]{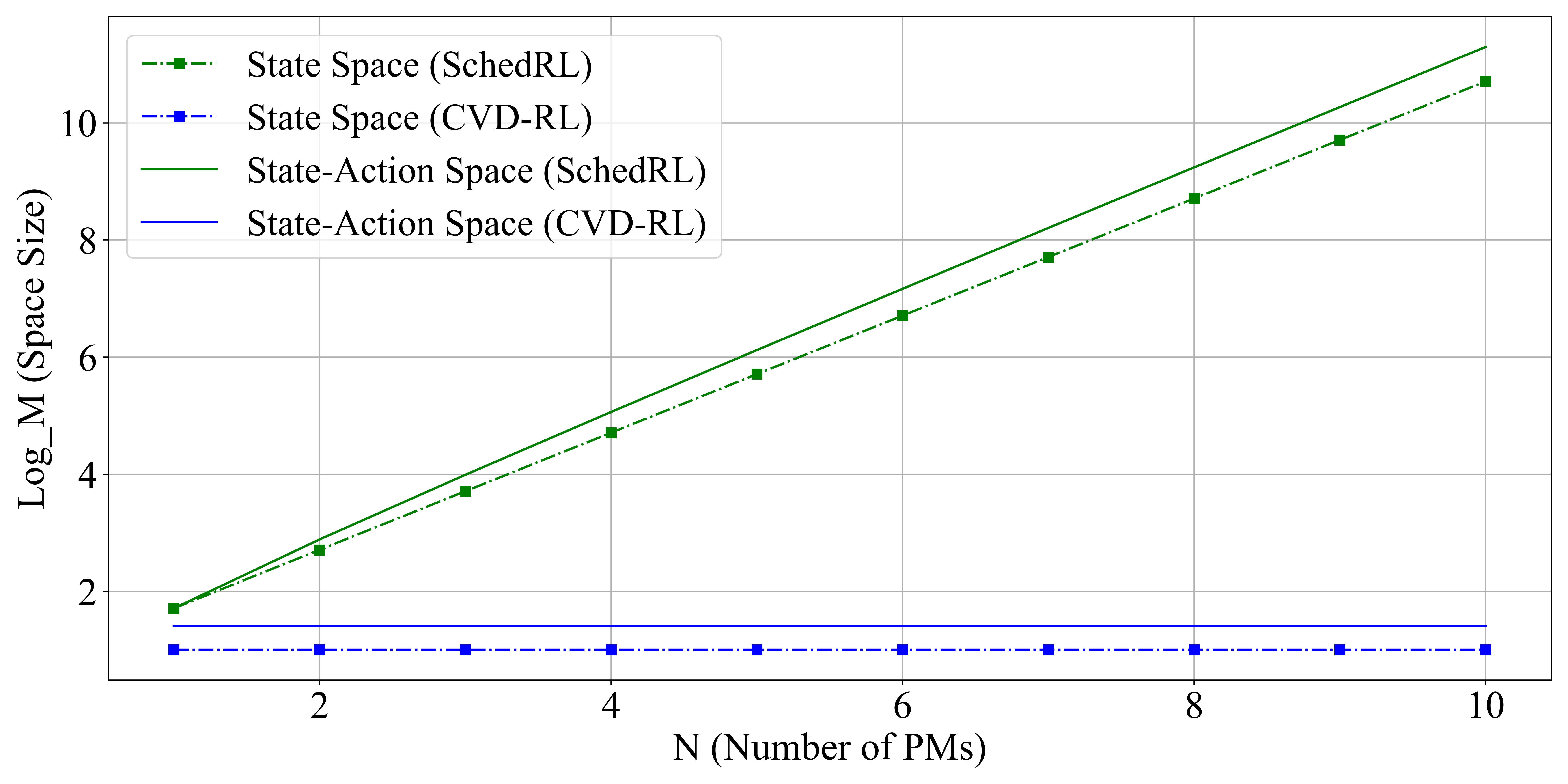}
    \end{minipage}
    \begin{minipage}[t]{0.45\textwidth}
        \centering
    \includegraphics[width=\textwidth]{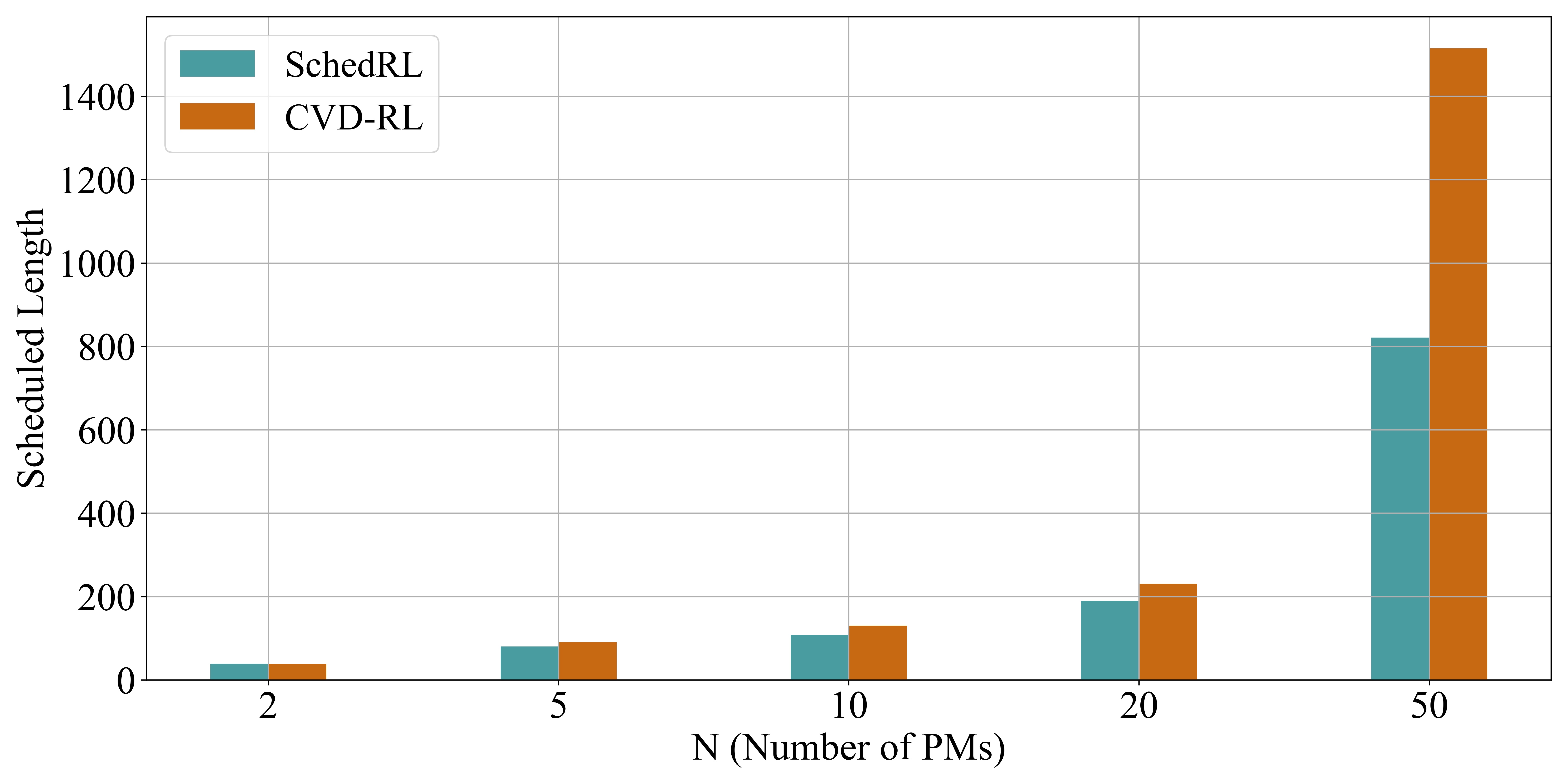}
    \end{minipage}
    \caption{{\bf{Top}}: The theoretical state-action spaces; {\bf{Bottom}}: The converged performances. The motivating cases for our method (CVD-RL). The left figure depicts the state-action spaces under different numbers of physical machines (PMs). For SchedRL, these spaces expand exponentially with an increasing number of PMs, in contrast to CVD-RL, where they remain constant. The right figure demonstrates the converged performance of various methods across different PM counts. Notably, the performance disparity between SchedRL and CVD-RL widens as the number of PMs grows.}
    \label{fig: scalability}
\end{figure}
This paper concentrates on advancing the scalability of RL for VMS within large-scale clusters. 
We identify two primary challenges: \textit{representation} and \textit{exploration}. 
As the number of PMs increases, the state space expands exponentially, as depicted in Figure \ref{fig: scalability}, presenting formidable representation challenges. 
The crucial task is enabling agents to extract meaningful and generalizable information from this vast state space. 
Additionally, with the expansion of the state-action space, as illustrated in Figure \ref{fig: scalability}, RL agents encounter significant \textit{exploration} challenges, making it impractical to explore the entire policy space comprehensively. 
Therefore, designing a strategy that facilitates high-quality exploration becomes essential.

\begin{figure*}[htb!]
    \centering
    \includegraphics[width=\textwidth]{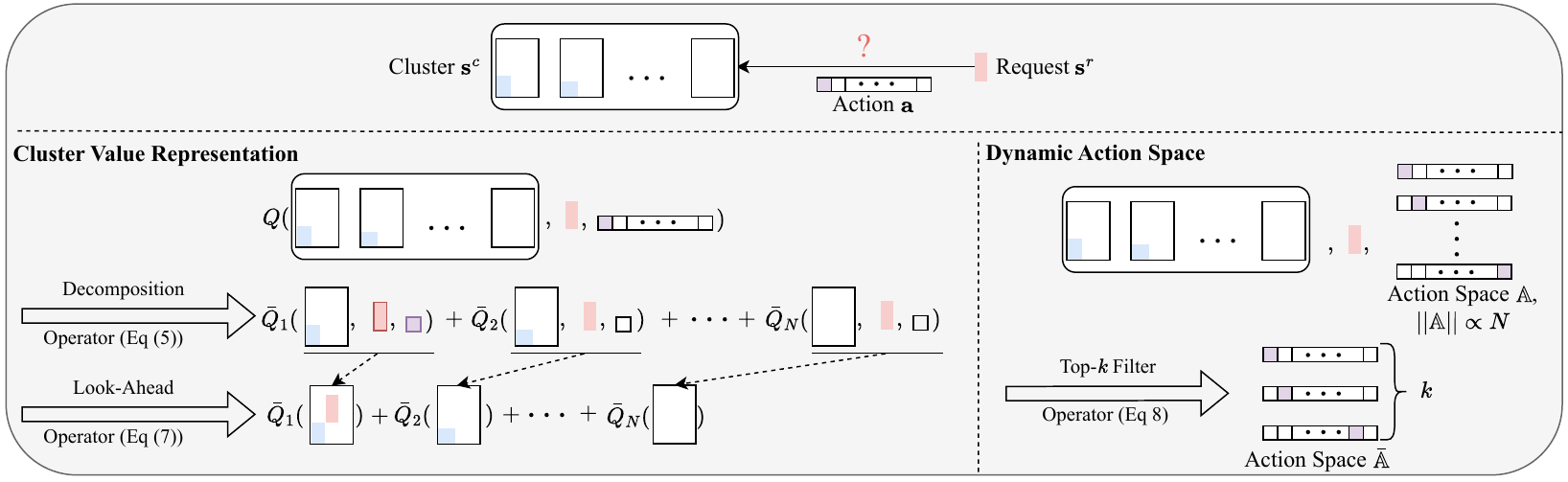}
    \caption{CVD-RL's scalability is enhanced by two key components: Cluster Value Representation and Dynamic Action Space Construction. The lower left segment illustrates the decomposition and look-ahead operator, which expresses the cluster's value as the aggregate of individual physical machines' values. Meanwhile, the bottom right details how the top-$k$ filter operator dynamically constructs an action space, effectively reducing the action space from linear growth to a fixed size of $k$.}
    \label{fig: schedmarl}
\end{figure*}

To address these challenges, we introduce the \textbf{C}luster \textbf{V}alue \textbf{D}ecomposition \textbf{R}einforcement \textbf{L}earning (\textbf{CVD-RL}) framework. 
The core idea of CVD-RL involves representing the value of the cluster through the values of individual PMs and generating a dynamic action space, as illustrated in Figure \ref{fig: schedmarl}. 
The framework incorporates three key operators: the decomposition operator, the look-ahead operator, and the filter operator. 
Both the decomposition and look-ahead operators are designed to tackle the representation challenge. 
Specifically, the decomposition operator breaks down the cluster's decision value into the sum of individual PMs' decision values, effectively mitigating the representation challenge and ensuring that popular scheduling strategies like Bestfit are included. 
The look-ahead operator further represents the value of each PM's decision as the value of the PM's future state, thereby obviating the need to extract complex data from VMs and actions. 
Addressing the exploration challenge, the filter operator employs a heuristic scheduling score function to create a dynamic yet high-quality action space for RL agents. 
Through these integrated operators, CVD-RL maintains a constant state and state-action space across varying numbers of PMs, as demonstrated in Figure \ref{fig: scalability}. 
Numerical results and ablation studies confirm the effectiveness of each operator. 
Notably, CVD-RL achieves significant improvements in CPU allocation efficiency, outperforming current SOTA methods. 
Its robustness is further validated in scenarios involving varying cluster usage patterns, larger clusters, and dynamic expansion conditions, marking a significant advancement in cloud computing resource management.

We summarize our contributions as follows:
(1) A dedicated and scalable DRL framework, {\bf{CVD-RL}}, is introduced for VMS.
The proposed CVD-RL reduces the stat-action space significantly and addresses the challenges of representation and exploration in large-scale clusters;
(2) Compared with existing RL-based VMS methods, CVD-RL can achieve superior scheduling performance on real-world data from Huawei Cloud up to $50$ PMs.
To emphasize, this can be considered the first successful application of an RL-based VMS at this scale in real-world datasets;
(3) The strategy obtained from CVD-RL exhibits remarkable generalizability across various configurations. It adapts effectively to different PM number, supports continual expansion in clusters, and is versatile in diverse take-over times.

\section{Preliminaries}

\subsection{Problem Statement}
\label{sec: problem}

In the realm of cloud computing, efficient resource management and VMS within a cluster environment are paramount. 
This paper delves into the algorithmic development for scheduling VMs across $N$ physical machines (PMs), each configured with \textbf{double Non-Uniform Memory Access} (NUMA) nodes, a setup \textit{prevalent in industrial contexts} due to its significant impact on resource allocation strategies \cite{rao2013optimizing,sheng2022learning}.

\begin{figure}[htb!]
    \centering
    \includegraphics[width=.8\linewidth]{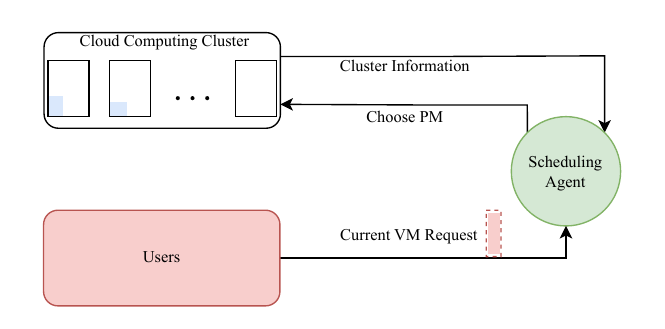}
    \caption{Overview of Virtual Machine Scheduling.}
    \label{fig: sched illu}
\end{figure}

To aid in understanding the VMS problem, Figure \ref{fig: sched illu} provides a sketch of the overall framework.
In this sketch, users continuously generate VM requests. 
Once a VM request is generated, the scheduling agent will receive it and monitor the current cluster information to decide which PM to allocate the VM. 
Specifically, each PM $i$ hosts two NUMA nodes, with the resource capacity of the $j$-th node represented as $c_{i,j}$. 
This model captures the multi-dimensional aspect of resource (CPU and memory) allocation challenges. VM requests are represented by tuples $\left\langle u, b \right\rangle$, encapsulating the required resources $u$ and the operation type $b$ (creation or release).
The scheduling process contains active allocations, where the scheduler decides the PM placement for incoming VM requests, and passive releases, which free up resources from previous allocations. 
VM requests are further classified as single-NUMA or double-NUMA based on their resource demands, with a special set $\mathbb{O}$ earmarked for double-NUMA requests that require resources from both NUMA nodes. 
The VMS problem then studies where each incoming VM request should be allocated to maximize the number of accommodated requests without overstepping the capacities.

\subsection{Problem Formulation}

We model the VMS problem as a MDP to accurately capture the dynamic of cloud computing environments. 
The state space \(\mathbb{S}\) consists of the cluster's status \(\boldsymbol{s}^c\) and the current request's status \(\boldsymbol{s}^v\). 
Specifically, \(\boldsymbol{s}^c\) describes the remaining resources across all \(N\) Physical Machines (PMs), and \(\boldsymbol{s}^v\) details the resources requested by the current VM. 
The variable \(s^c_{i,j}\) represents the remaining resources of the \(j\)-th Non-Uniform Memory Access (NUMA) node in the \(i\)-th PM.

The action space \(\mathbb{A}\) is a \(2N\)-dimensional one-hot vector that identifies the target PM and NUMA node for allocation. 
For each $\boldsymbol{a}\in \mathbb{A}$, its index refers to the target NUMA and PM for single-NUMA requests while refers to the target PM for double-NUMA requests. 
Suppose a cluster with $2$ PMs, each comprising $2$ NUMA nodes. 
The single-NUMA action space contains $4$ possible one-hot vectors. 
For instance, 
\begin{itemize}[leftmargin=*]
    \item $[1, 0, 0, 0]$ indicates assigning a single-NUMA request to the $1$st NUMA node of the $1$st PM, and
    \item $[0, 0, 0, 1]$ indicates assigning a single-NUMA request to the $2$nd NUMA node of the $2$nd PM.
\end{itemize} 
For a single-NUMA VM, one of these $4$ entries is selected. 
For a double-NUMA VM, it is split evenly across both NUMA nodes of the same PM. 
Hence, $[1, 0, 0, 0]$ (or $[0, 1, 0, 0]$) implies a double-NUMA request is allocated to the two NUMA nodes of the $1$st PM, while $[0, 0, 1, 0]$ (or $[0, 0, 0, 1]$) indicates the $2$nd PM.
Our reward function \(R\) adopts the \(\delta\)-reward mechanism proposed by \cite{sheng2022learning}, which evaluates the immediate impact of scheduling decisions, thereby enhancing learning efficiency in VMS tasks.

\begin{figure}[htb!]
    \centering
    \includegraphics[width=.8\linewidth]{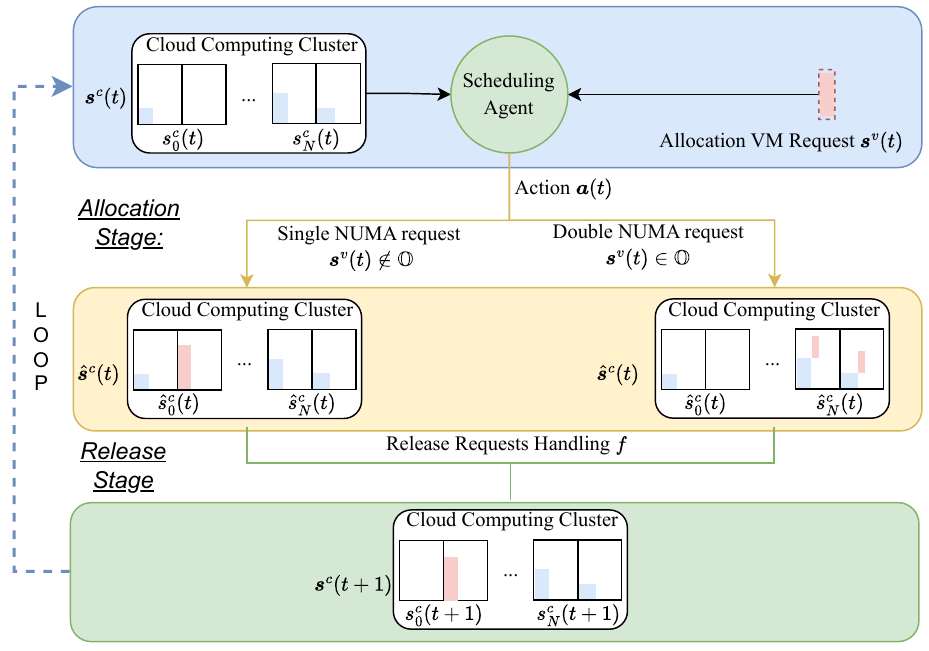}
    \caption{VMS dynamics. The process begins with the initial cluster status \(\boldsymbol{s}^c(t)\) and the current allocation VM request \(\boldsymbol{s}^v(t)\). The scheduling agent selects a scheduling action \(\boldsymbol{a}(t)\) accordingly, causing the cluster to transition to \(\hat{\boldsymbol{s}}^c(t)\) based on the action and the double-NUMA set \(\mathbb{O}\). Afterward, the system handles release requests through a function \(f\) and transitions to \(\boldsymbol{s}^c(t+1)\), continuing to handle the next VM request.}
    \label{fig: schedform}
\end{figure}

The transition dynamics are articulated through a series of equations. 
We split the transition into two stages: allocation and release, as shown in Fig.~\ref{fig: schedform}. 
In the allocation stage, the scheduling agent selects an action \(\boldsymbol{a}\) for the current allocation request \(\boldsymbol{s}^v(t)\) based on the current cluster status \(\boldsymbol{s}^c(t)\). 
The objective is to discover a policy \(\pi\) that maximizes the expected cumulative reward, thereby optimizing resource utilization in response to fluctuating VM requests. 
The cluster status then transitions to \(\hat{\boldsymbol{s}}^c\), reflecting the resource allocation.
In the release stage, the cluster handles subsequent release requests until the next allocation request arrives. 
Since the release requests specify their locations, the scheduling agent does not need to make decisions during this process. 
We denote this release process as controlled by function \(f\).
The overall problem can be defined as:
\begin{align}
& \max_\pi \quad J(\pi)=\mathbb{E}_{\boldsymbol{s}(0)} \left[\textstyle{\sum_{t=0}^T \gamma^t R\left(\boldsymbol{s}(t), \boldsymbol{a}(t)\right)}\right],\;s.t.,\nonumber\\
& \boldsymbol{a}(t) = \pi\left(\cdot \mid \boldsymbol{s}(t)\right), \quad \ell=\arg\max(\boldsymbol{a}(t)), \nonumber \\
& \hat{s}^c_i(t+1) = \boldsymbol{s}^c_i(t), \quad \forall i \neq \lfloor{\ell/2}\rfloor, \label{eq:constraint_nonselected}\\
& \hat{s}_{i}^c(t+1) = s_i^c(t) - \boldsymbol{s}^v(t) \odot \ell, \quad \text{if } i = \lfloor{\ell/2}\rfloor, \label{eq:constraint_selected} \\
& \hat{s}_{i}^c(t+1) \geq 0 , \quad \forall i<N, \label{eq:non-neg}\\
& {s}_{i}^c(t+1) = f(\hat{s}_i^c(t+1)), \quad \forall i<N, \label{eq:release}
\end{align}
where \(\ell\) is the action \(\boldsymbol{a}\) represented as an integer. 
Equations \eqref{eq:constraint_nonselected} and \eqref{eq:constraint_selected} define how a PM's status transitions after handling the VM request according to action \(\boldsymbol{a}(t)\). 
Equation \eqref{eq:non-neg} ensures that no PM's resources are negative after the allocation stage. 
Equation \eqref{eq:release} describes the release stage. 

To be concise, we introduce the operator \(\odot\) to denote the allocation handling process, defined as:
\begin{equation}
\!\!\!\!\!\!\boldsymbol{s}^v(t) \odot \ell  =
\begin{cases}
[\boldsymbol{s}^v(t),\vec{0}], & \text{if } \ell\%2=0 \text{ and } \boldsymbol{s}^v(t) \notin \mathbb{O}, \\
[\vec{0}, \boldsymbol{s}^v(t)], & \text{if } \ell\%2=1 \text{ and } \boldsymbol{s}^v(t) \notin \mathbb{O}, \\
[\frac{\boldsymbol{s}^v(t)}{2}, \frac{\boldsymbol{s}^v(t)}{2}], & \text{if } \boldsymbol{s}^v(t) \in \mathbb{O},
\end{cases}
\label{eq: hat_s}
\end{equation}
where \(\vec{0}\) denotes zero resources in a NUMA. 
\section{Double DQN For Scheduling}
\label{sec: DDQN}
To address the Markov Decision Process (MDP) delineated in Section~\ref{sec: problem}, this study employs Double Deep Q-Networks (Double DQN) as the foundational framework. 
We utilize $Q(s, a; \theta)$ to estimate the expected cumulative reward of scheduling request $s^v$ to physical machine (PM) $a$ within the cluster state $s^c$, where $\theta$ represents the neural network parameters. 
The decision-making policy is expressed as:
\begin{equation}
    \pi(\cdot \mid \boldsymbol{s}) = \arg\max_{\boldsymbol{a} \in \mathbb{A}} Q(\boldsymbol{s}, \boldsymbol{a}; \theta). \nonumber
\end{equation}
In addition, the target network $Q(\boldsymbol{s}, \boldsymbol{a}; \theta')$ is maintained to enhance training stability.
For each transition tuple $\langle \boldsymbol{s}(t), \boldsymbol{a}(t), r(t), \boldsymbol{s}(t+1) \rangle$, the network parameters are updated as follows:
\begin{equation}
    \begin{aligned}
        &\min_\theta \big\| Q\left(\boldsymbol{s}(t), \boldsymbol{a}; \theta\right) - Y(t) \big\|^2_2,\;\text{where} \\
        Y(t) = r(t) & + \gamma Q \left(\boldsymbol{s}(t+1), \arg\max_{\boldsymbol{a} \in \mathbb{A}} Q\left(\boldsymbol{s}(t+1), \boldsymbol{a}; \theta\right); \theta'\right),\nonumber
    \end{aligned}
\end{equation} where $\gamma$ is the discount factor.
The target network employs a soft updating mechanism:
\begin{equation}
    \theta' = \lambda \theta + (1 - \lambda) \theta',
    \label{eq: soft-upd}
\end{equation} with $\lambda$ being the soft update rate.
Given a cluster with $N$ PMs, where the PM and VM space size are $M$ and $L$, the state space for Double DQN is $LM^N$ and the action space is $2N$. 
This scale presents substantial challenges in terms of representation and exploration, particularly as the number of PMs increases, as illustrated in Fig.~\ref{fig: scalability}.

\section{The Scalable Algorithm Framework}

This section delineates the CVD-RL method, crafted specifically for VMS. 
It proposes a novel cluster value representation and the dynamic action space construction technique to improve the scalability of the RL agent in VMS.  
The cluster value representation is built by a decomposition operator and a look-ahead operator while the dynamic action space is constructed by the top-$k$ filter operator. 
We commence by delineating three innovative operators integral to CVD-RL, followed by an overview of the complete algorithm.

\subsection{Decomposition Operator}
The cluster status $\boldsymbol{s}^c$ makes a key contribution to the exponentially increasing state-action space as analyzed in \ref{sec: DDQN}. 
To address this issue, we propose the decomposition operator. 

We first assume that the cluster's state-action value function can be approximately decomposed into the addition of the value across PMs. 
This assumption aligns with common practices in multi-agent RL \cite{sunehag2017value} and cloud resource estimation \cite{sheng2023learning}.
With this assumption, we have 
\begin{equation}\label{eq: decomposition}
\begin{aligned}
 Q \left( \boldsymbol{s}^c(t), \boldsymbol{s}^v(t), \boldsymbol{a}; \theta \right) \approx \sum_i \bar{Q}_i \left( s^c_{i}(t), \boldsymbol{s}^v(t), \bar{a}_i; \theta_i \right),
\end{aligned}
\end{equation}where $\bar{a}_i= [{a}_{2i}, {a}_{2i+1}]$ and $\bar{Q}_i$ is the value function of PM $i$. Now the $\bar{Q}_i$ only depends on the physical machine $i$'s state and each PM only perceives state space as $LM$.  
Further, we employ parameter sharing across PMs as $\theta_{0}=\theta_2=\dots=\theta_{N-1}=\bar{\theta}$.
The state-action space thus scales linearly with the number of PMs and the sample efficiency gets improved significantly. 
With such a representation, we derive the scheduling policy as below. 
\begin{equation}
\begin{aligned}
\pi \left(\boldsymbol{a}\, \big| \, \boldsymbol{s}; \bar{\theta} \right) &:= \arg\max_{\boldsymbol{a}\in \mathbb{A}} \sum_i^N \bar{Q}_i \big(s_i^c(t), \boldsymbol{s}^v(t), \bar{a}_i(t); \bar{\theta} \big).
\label{eq: decomp_pi}
\end{aligned}
\end{equation}
Due to the action space $\mathbb{A}$ only containing one-hot actions, the $\arg\max$ can be further taken on the level of PM. 
For each PM, it can select the action in $\bar{\mathbb{A}}_i$ which includes not schedule to this PM $[0,0]$, schedule to the first NUMA $[1,0]$ and to the second NUMA $[0, 1]$. 
We denote the $[0, 0]$ action as $\hat{a}_i$. 
Thus each PM evaluates its action benefits as: 
$$\triangle_i(\bar{a}_i) = \bar{Q}_i(s_i^c(t), \boldsymbol{s}^v(t), \bar{a}_i; \bar{\theta}) - \bar{Q}_i(s_i^c(t), \boldsymbol{s}^v(t), \hat{a}_i; \bar{\theta}).$$
Thus \eqref{eq: decomp_pi} can be decomposed as two steps. For the first step, each PM evaluates its action benefits and the highest benefits one obtains the scheduled identity $i^*$: $i^*=\arg\max_i \max_{\bar{a}_i}\triangle_i(\bar{a}_i) $. In the second step, each PM selects its own action as:
$$
\bar{a}_{i}^* = \left\{ \begin{array}{l}
     [0,0],\ \text{ if }i\neq {i}^*,\\
     \arg\max\limits_{\bar{a}_{i}\in\bar{\mathbb{A}}_i} \triangle_i(\bar{a}_i), \text{otherwise}
    \end{array}\right.
$$
Such action selection aligns to \eqref{eq: decomp_pi} which forces the joint action to be one-hot and the highest score one is selected. 

\paragraph{Remark 1}
The decomposition operator does not make the policy space too strict. 
For example, the Best-fit policy is in our policy space. 
This indicates that the assumption does not sacrifice commonly used policy during learning.
For the proof, we refer readers to the Appendix.

\paragraph{Remark 2} 
The decomposition operator is similar to MARL. 
They all make decomposition for the centralized value function. 
However, different from many MARL methods~\cite{sunehag2017value}, each PM's action selection is not independent where only one PM's NUMA is chosen. 
Thus the centralized action selection is needed. 

\paragraph{Remark 3}
Though our principal application centers on online vector bin-packing for cloud resource allocation, the underlying ideas can transfer to other multi-dimensional scheduling problems. 
Specifically, the decomposition operator (for large state spaces) and top-filtering operator (for large action spaces that may be pruned via domain heuristics) are quite general, whenever one needs to systematically reduce an otherwise enormous state–action space.

\subsection{Look-Ahead Operator}
Due to the impact of scheduling action being straightforward to predict, we propose the look-ahead operator to reduce the state space further. 
As shown in \eqref{eq:constraint_nonselected} and \eqref{eq:constraint_selected}, only resources in selected PM will be influenced during the allocation stage.
To exploit this knowledge, we represent the state action value function as 
\begin{equation}
\bar{Q}_i \left( s^c_{i}(t), \boldsymbol{s}^v_t, \bar{a}_i; \bar{\theta} \right) = \bar{Q}_i \left( \hat{s}^c_i(t+1); \bar{\theta} \right),\ \forall i<N,
\label{eq: look-ahead}
\end{equation}
where $\hat{s}^c_i(t+1)$ is obtained from \eqref{eq:constraint_nonselected} and \eqref{eq:constraint_selected}.
The intermediate states $\hat{s}(\cdot)$ do not eliminate the reward term; rather, we incorporate it into the next-step value.
For convenience, we abuse $\bar{Q}_i$ that allows it to accept both $(s^c_{i}(t), \boldsymbol{s}^v_t, \bar{a}_i)$ and $\hat{s}^c_i(t+1)$. 
While in practice, each $\bar{Q}_i \left( s^c_{i}(t), \boldsymbol{s}^v_t, \bar{a}_i; \bar{\theta} \right)$ will be calculated as \eqref{eq: look-ahead}.
This operator transforms the PM-VM-action value estimation into the allocated PM's value estimation. 
Thus, the state space is further reduced to $M$, and the state-action space to $MN$, making our framework adaptable to various VM configurations.


\subsection{Top-k Filter Operator}
To address the exploration challenge, especially with an increasing number of PMs (where the action space grows linearly), we introduce the Top-$k$ Filter Operator. 
This operator utilizes domain knowledge to construct a more efficient exploration space. 
While it may be challenging for domain experts to specify the most promising actions, heuristic schedulers can provide useful guidance.

W.l.o.g., we consider a heuristic score function $g$ that scores each action $\boldsymbol{a}$ under each state $\boldsymbol{s}$. 
The Top-$k$ filter operator then suggests a subset of the action space $\hat{\mathbb{A}} \subset \mathbb{A}$ as follows:
\begin{equation}
\hat{\mathbb{A}} =\underset{\boldsymbol{a} \in \mathbb{A}}{\mathrm{arg\ top}\text{-}k}\ g(\boldsymbol{a}, \boldsymbol{s}).
\label{eq: topk}
\end{equation}
This action space is dynamically constructed and we let the RL agent interact within this action space.
Subsequently, the RL scheduler selects actions using $\hat{\mathbb{A}}$. 
The policy with top-$k$ filter operator then can be derived as:
\begin{equation}
\begin{aligned}
\pi(\boldsymbol{a}\mid \boldsymbol{s}(t); \bar{\theta}) = \arg\max_{\boldsymbol{a} \in \hat{\mathbb{A}}} \sum_i^N \bar{Q}_i(\hat{s}_i^c(t+1); \bar{\theta}),
\end{aligned}
\label{eq: greedy}
\end{equation}
where $\hat{s}_i^c(t+1)$ is obtained as \eqref{eq:constraint_nonselected} and \eqref{eq:constraint_selected}. 
This equation is derived from applying decomposition and look-ahead operators. 
This approach effectively reduces the action space from $2N$ to the size of $k$, which remains constant despite an increase in the number of PMs.

The formulation of the score function \(g\) and the choice of \(k\) are pivotal elements that significantly influence the learning process, necessitating a balanced consideration of both diversity and quality within the scheduling framework. 
In this context, we opt for \(k=5\) as the default setting, drawing upon the Best-Fit heuristic and our Internal-Scheduler developed for this purpose to delineate the score function. 
Specifically, the score function \(g(\boldsymbol{a}, \boldsymbol{s})\) assigns a value of \(1\) to actions that are ranked within the top-2 by the Best-Fit heuristic or within the top-3 by the Internal-Scheduler, with all other actions receiving a score of \(0\). 
This approach is mathematically represented as follows:
\begin{equation}
    g(\boldsymbol{a}, \boldsymbol{s}) = \begin{cases} 
    1, & \text{if } \boldsymbol{a} \in \text{Top-2}_{\text{BF}}(\boldsymbol{s}) \text{ or } \boldsymbol{a} \in \text{Top-3}_{\text{IS}}(\boldsymbol{s}), \\
    0, & \text{otherwise},
    \end{cases}
    \nonumber
\end{equation} where \(\text{Top-2}_{\text{BF}}(\boldsymbol{s})\) denotes the set of actions ranked as the top-2 by the Best-Fit heuristic for state \(\boldsymbol{s}\), and \(\text{Top-3}_{\text{IS}}(\boldsymbol{s})\) represents the set of actions within the top-3 as determined by the Internal-Scheduler for the same state.

\subsection{Algorithm Framework}
The CVD-RL algorithm provides a systematic approach to VMS, leveraging RL to optimize decisions in dynamic environments. 
Algorithm~\ref{alg: CVD-RL} outlines the procedure. 
At each decision step, the agent first obtains a refined action space with Top-$k$ Filter Operator. 
Then it further applies the decomposition operator and look-ahead operator to construct policy $\pi$ with \eqref{eq: greedy}. 
Taking $\epsilon-$greedy as the exploration scheme in the refined action space, the agent collects a number of transitions and stores them into a replay buffer. 
As for learning procedure, the agent update $\bar{\theta}$ as 
\begin{align}\label{eq: loss}
    &\min_{\bar{\theta}} \big\|\sum_i^N \bar{Q}_i(\hat{s}_i^c(t+1); \bar{\theta})-Y(t)\big\|_2^2,\;\text{where}\\
    Y(t)&=r(t)+\gamma \sum_i^N \bar{Q}_i \left( s_i^c({t+1}), \pi(\boldsymbol{a}|\boldsymbol{s}(t+1);\bar{\theta}); \bar{\theta}' \right).\notag
\end{align}

\begin{algorithm}[htb!]
\caption{CVD-RL for VMS}
\label{alg: CVD-RL}
\begin{algorithmic}[1]
\STATE \textbf{Initialization:} Initialize $Q$-network parameters $\bar{\theta}$, target $Q$-network parameters $\bar{\theta}'$, score function $g$, discount factor $\gamma$, and set maximum epochs $E$;
\FOR{each episode $= 1$ to $E$}
    \STATE Begin with a sample sequence and initial state $\boldsymbol{s}(0)$;
    \FOR{each decision step $t=0$ to $T$}
        \WHILE{$d(t) \neq \text{False}$}
            \STATE Apply Top-$k$ Filter Operator to obtain $\bar{\mathbb{A}}$ at state $\boldsymbol{s}(t)$ using score function $g$ with \eqref{eq: topk};
            \STATE Apply Decomposition and Look-Ahead Operator and obtain policy $\pi$ as \eqref{eq: greedy};
            \STATE Adopt $\epsilon$-greedy strategy: Choose a random action from $\bar{\mathbb{A}}$ with probability $\epsilon$, or use the policy $\pi$ for action selection with probability $1-\epsilon$;
            \STATE Execute the action, observe rewards $r(t)$, and transition to new state $\boldsymbol{s}(t+1)$;
            \STATE Store transition $(\boldsymbol{s}(t), \boldsymbol{a}(t), r(t), \boldsymbol{s}(t+1))$;
            \STATE Sample a batch from the replay buffer and update $\theta$ by minimizing the loss in \eqref{eq: loss};
            \STATE Periodically update $\bar{\theta}'$ with $\bar{\theta}$ using soft update rule in \eqref{eq: soft-upd}.
        \ENDWHILE
    \ENDFOR
\ENDFOR
\end{algorithmic}
\end{algorithm}

\paragraph{Remark 4}

one can view our problem as an online bin-packing scenario with uncertain VM arrivals and resource demands. 
A pure constraint programming (CP) approach (even via a stochastic CP framework) might look like:
$$
\min \sum_{t=1}^{T} \sum_{i=1}^{N} c\bigl(x_{t,i}\bigr)y_{t,i}, 
\quad 
\text{s.t.}
\quad
P\bigl(g_k(x_{t,i}, \xi) \le 0\bigr) \ge \alpha_k, k=1,\dots,K,
$$ where $\xi$ denotes the stochastic elements (e.g., future arrival rates or resource fluctuation). 
In principle, such a formulation could capture the constraints, but significant barriers prevent widespread deployment at large scale:
\begin{itemize}[leftmargin=*]
    \item \emph{Non-convex probability constraints}. Computing or approximating $P(\cdot)\ge\alpha_k$ can become highly complex, especially if the distribution of $\xi$ is unknown or time-varying.
    \item \emph{Time sensitivity}. In production environments, scheduling decisions often must be made in milliseconds. Solving even moderately sized CP models repeatedly in such time frames can be infeasible.
    \item \emph{Rolling horizon demands}. Real systems require continual re-optimization as new VMs arrive over time. This frequent re-solving of stochastic CP can become computationally prohibitive.
\end{itemize}
By contrast, a reinforcement learning policy can learn from experience to approximate near-optimal decisions under uncertainty, making split-second allocations without repeatedly invoking a solver.

\section{Experiments}

This section outlines the simulation environment, test scenarios, baseline models, and metrics for performance evaluation. 
We then present a comparative analysis of our proposed model against these baselines under various conditions.

\subsection{The Environment and Scenarios}

Our experiments utilize VMAgent~\cite{sheng2022vmagent}, a sophisticated simulation environment that integrates real-world cloud scheduling data from Huawei Cloud. 
This setup provides a realistic approximation of cloud computing environments for scenario modeling.

\begin{figure}[ht]
    \centering
    \includegraphics[width=0.6\textwidth]{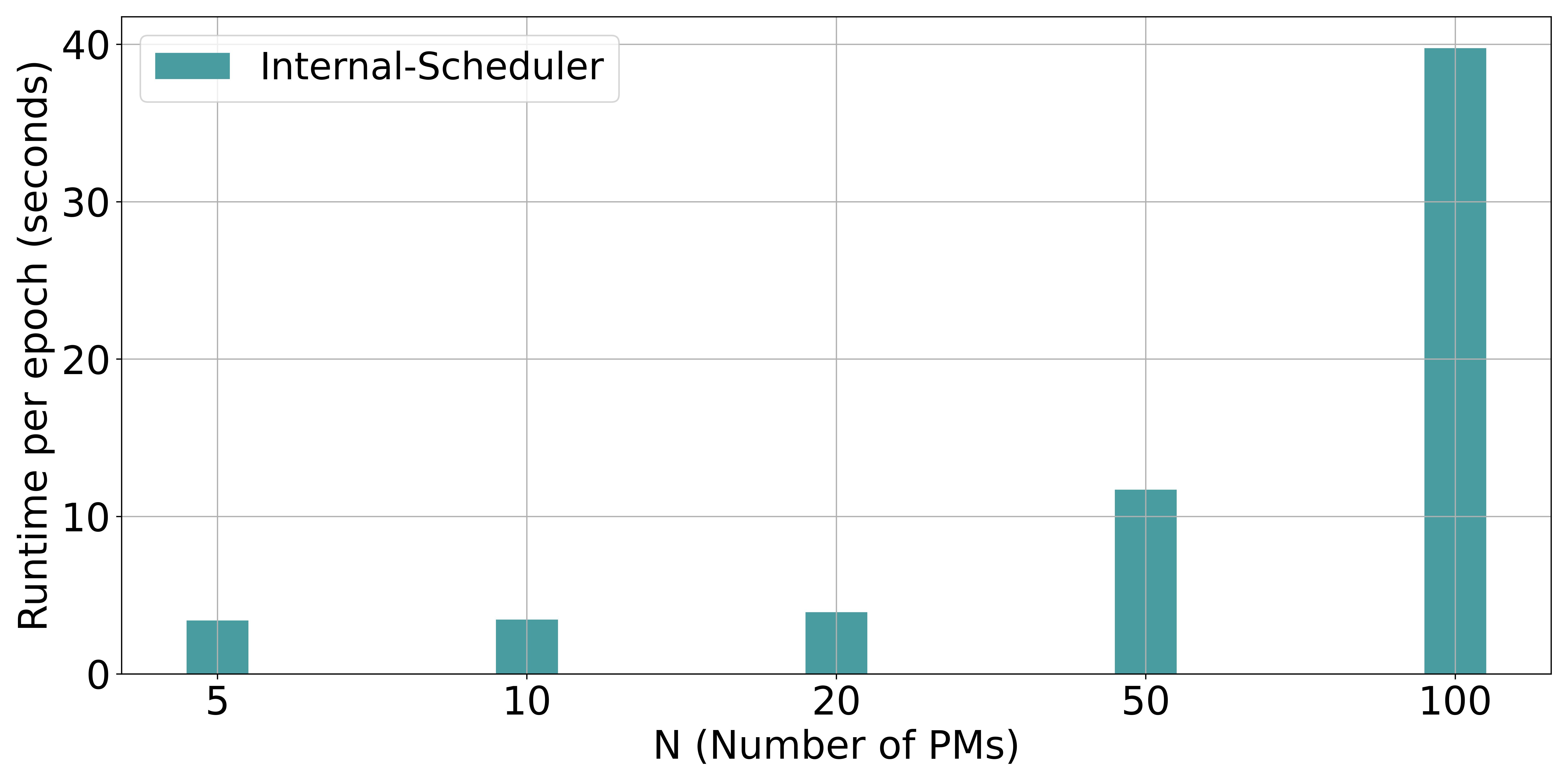}
    \caption{Runtime comparisons on Non-Expansion scenarios with increasing numbers of PMs, highlighting the trend of increased sampling time as the cluster size grows.}
    \label{fig: runtime}
\end{figure}

We focus on two key aspects in our test scenarios: the warm start ratio (\(\rho_{ws}\)) and scenario scalability. 
The warm start ratio, defined as the CPU usage threshold that triggers the scheduling process, is crucial in determining the distribution of physical machines (PMs). 
For scenario scalability, the number of PMs (\(N_{PMs}\)) is the key factor. 
We consider two settings: Non-Expansion and Expansion. 
In the Non-Expansion setting, the environment terminates when the current VM request cannot be handled. 
In the Expansion setting, the cluster adds a few PMs to the cluster each time it cannot handle a request until the maximum number of PMs is reached. 
These scenarios are common in cloud servicing.

Before specifying our scenarios, we first estimate runtime costs using the state-of-the-art scheduler, Internal Scheduler\footnote{Internal Scheduler is a best-fit variant which incorporates domain-specific rules.}, to calculate the average running time for a single episode with \(50\%\) warm start and various numbers of PMs. 
A significant portion of our experiment costs is attributed to sampling time, which increases with the number of PMs. 
The trend, depicted in Fig.~\ref{fig: runtime}, illustrates that the runtime for each epoch escalates significantly as the number of PMs increases, underlining the computational complexity associated with larger clusters. 
For example, the sampling time of \(3000\) epochs with \(100\) PMs requires roughly \(33\) hours, compared to \(9\) hours for \(50\) PMs.

For the training scenario, we choose a challenging setting\footnote{The number of PMs is also a key factor in the applicability testing. 
A cluster size of \(50\) PMs is considered challenging for RL methods to converge \cite{raven}. Many data centers are partitioned into clusters of tens or hundreds of PMs, which is consistent with testing at a $50$–$100$ PM scale. While commercial clouds can exceed thousands of PMs per cluster, heuristic solutions dominate at that size. Our method shows that an RL-based approach can outperform such heuristics in moderate clusters, possibly opening avenues to scale further with additional research.} with a \(50\%\) warm start ratio and \(50\) PMs in Non-Expansion mode to reduce the Markov Decision Process (MDP) horizon and improve sampling efficiency. 
For the testing scenario, we examine the generalization ability of the policy learned through trial and error. 
We test warm start ratios of \(0\%\), \(30\%\), \(40\%\), \(50\%\), and \(60\%\) to explore their impact on deployment constraints and generalization capabilities. 
We also assess the policy's applicability to larger numbers of PMs and dynamic PM counts by constructing a Non-Expansion scenario with \(100\) PMs and an Expansion scenario. 
The Expansion scenario starts with \(50\) PMs and incrementally adds \(10\) PMs each time the cluster's capacity is exceeded, up to a total of \(110\) PMs.

To facilitate result reproduction, we provide hyperparameters in Table~\ref{tb: hyper}. 
For the neural network architecture, we employ a 6-layer Multi-Layer Perceptron (MLP) with ReLU activation \cite{relu}. 

\begin{table}[htbp!]
  \centering
  \caption{Hyperparameters of Our CVD-RL Model}
  \label{tb: hyper}
  \begin{tabular}{cc|cc}
    \hline
    \textbf{Parameter} & \textbf{Value} & \textbf{Parameter} & \textbf{Value} \\
    \hline
    Learning Rate     & \(5 \times 10^{-4}\) & Epochs            & 3000  \\
    \(\gamma\)        & 0.75                 & \(\epsilon\)      & 0.1 \\
    Batch Size        & 2048                 & Buffer Capacity   & \(10^5\) \\
    Optimizer         & Adam                 &  \(\tau\)          & 0.01 \\
    \hline
  \end{tabular}
\end{table}

\subsection{Baseline Models and Evaluation Metrics}

We evaluate our model against six baseline algorithms: First-Fit, Best-Fit, Internal-Scheduler, SchedRL~\cite{sheng2022learning}, PADRL~\cite{zeng2022adaptive} and Hindsight Learning (HL)~\cite{sinclair2023hindsight}. 
First-Fit and Best-Fit are traditional heuristics used in cloud computing, with First-Fit allocating the first PM that can accommodate a request, and Best-Fit prioritizing PMs with the least remaining CPU resources. 
The Internal-Scheduler is an industrial standard from our organization. SchedRL represents a SOTA RL-based scheduler for comparison. 
PADRL represents the powerful RL scheduler in VM Rescheduling. 
HL is the recent learning-based VM scheduler.

Our evaluation employs three metrics from VMAgent:

\noindent - {\em{Scheduled Length}}: The total number of requests processed by the cluster from start to finish. A longer scheduled length indicates a more effective scheduling algorithm;

\noindent - {\em{Average CPU Utilization}}: The mean CPU utilization across all test scenarios. Higher average utilization reflects greater resource efficiency;

\noindent - {\em{Income}}: Derived from VM request durations and hourly rates from Huawei Cloud, this metric approximates the revenue potential of the scheduling algorithm. Higher-income suggests superior performance.

\begin{table*}[!htb]
    \centering
    \caption{Comparative Performance Analysis of First-Fit, Best-Fit, Internal-Scheduler, SchedRL, PADRL, HL, and CVD-RL Strategies in Non-Expansion Scenario with 50 Physical Machines (PMs). The table includes performance metrics such as Length, Income, and CPU Allocation (CpuAllo) across various warm start ratios ($\rho_{ws}$).}
    \label{tb: performance-50}
    \resizebox{\textwidth}{!}{%
    \begin{tabular}{c|c|c|c|c|c|c|c|c}
        \hline
        $\rho_{ws}$ & Metric & First-Fit & Best-Fit & Internal-Scheduler & SchedRL & PADRL & HL & CVD-RL \\ 
        \hline
        \multirow{3}{*}{0\%} & Length & 1282.3 & 1386.8 & 1398.6 & 1024.7 (±9.8) & 1008.74 (± 37.37) & 1032.7 (± 8.1) & \textbf{1464.6} (±23.8) \\ 
        ~ & Income & 6942.1 & 7880.5 & 8191.5 & 4905.5 (±110.3) & 4690.24 (± 294.01) & 5057.87 (± 90.21) & \textbf{8680.2} (±119.7) \\ 
        ~ & CpuAllo & 39.16\% & 41.7\% & 41.6\% & 34.2 (±0.2)\% & 33.78 (± 19.01)\% & 34.14 (± 19.49)\% & \textbf{43.2} (±0.4)\% \\ 
        \hline
        \multirow{3}{*}{30\%} & Length & 1050.4 & 1149.7 & 1144.2 & 861.6 (±24.1) & 830.99 (± 10.12) & 843.64 (± 16.21) & \textbf{1224.3} (±37.5) \\ 
        ~ & Income & 6178.2 & 7157.4 & 7751.7 & 4882.4 (±218.0) & 4562.01 (± 251.1) & 4731.59 (± 198.74) & \textbf{7883.6} (±317.8) \\ 
        ~ & CpuAllo & 48.18\% & 50.4\% & 50.6\% & 45.2 (±0.4)\% & 44.57 (± 25.63)\% & 45.09 (± 25.86)\% & \textbf{51.4} (±0.2)\% \\ 
        \hline
        \multirow{3}{*}{40\%} & Length & 910.5 & 1010.1 & 1004.5 & 721.8 (±3.0) & 735.94 (± 28.18) & 744.44 (± 7.43) & \textbf{1061.9} (±8.5) \\ 
        ~ & Income & 5344.7 & 6244.2 & 6428.6 & 3933.5 (±45.8) & 4024.7 (± 126.33) & 4145.16 (± 160.5) & \textbf{6789.3} (±267.8) \\ 
        ~ & CpuAllo & 51.42\% & 53.9\% & 53.8\% & 48.3 (±0.3)\% & 48.29 (± 27.95)\% & 48.83 (± 27.85)\% & \textbf{54.5} (±0.3)\% \\ 
        \hline
        \multirow{3}{*}{50\%} & Length & 806.9 & 852.8 & 844.0 & 620.3 (±18.0) & 627.09 (± 36.55) & 642.33 (± 2) & \textbf{939.3} (±23.8) \\ 
        ~ & Income & 4805.3 & 5295.4 & 5112.9 & 3329.8 (±219.0) & 3529.06 (± 207.7) & 3555.2 (± 119.69) & \textbf{6273.4} (±5.0) \\ 
        ~ & CpuAllo & 56.40\% & 57.4\% & 57.5\% & 52.9 (±0.2)\% & 52.97 (± 30.28)\% & 53.51 (± 30.62)\% & \textbf{58.4} (±0.2)\% \\ 
        \hline
        \multirow{3}{*}{60\%} & Length & 593.3 & 669.2 & 661.9 & 449.4 (±17.5) & 455.09 (± 25.53) & 474.22 (± 14.19) & \textbf{712.4} (±3.2) \\ 
        ~ & Income & 3354.9 & 3947.8 & 4391.4 & 2280.4 (±139.7) & 2237.35 (± 175.95) & 2470.33 (± 91.76) & \textbf{4441.9} (±191.9) \\ 
        ~ & CpuAllo & 62.29\% & 63.2\% & 63.3\% & 59.9 (±0.2)\% & 59.67 (± 34.12)\% & 60.36 (± 34.36)\% & \textbf{63.7} (±0.0)\% \\ 
        \hline
    \end{tabular}%
    }
\end{table*}

\begin{table}[!htb]
    \centering
    \caption{Performance comparison of First-Fit, Best-Fit, Internal-Scheduler, and CVD-RL strategies across Non-Expansion scenarios with 100 PMs. The CVD-RL's policy is trained in 50 PMs and tested in the Non-Expansion scenario with 100 PMs. Due to SchedRL, PADRL, and HL's trained policy can not be applied to scenarios with dynamic numbers of PMs, we omit them in the table. The table includes performance metrics such as Length, Income, and CPU Allocation (CpuAllo) across various warm start ratios ($\rho_{ws}$).}
    \label{tb: performance-100}
    \resizebox{.8\linewidth}{!}{%
    \begin{tabular}{c|c|c|c|c|c}
        \hline
        $\rho_{ws}$ & Metric & First-Fit & Best-Fit & Internal-Scheduler & CVD-RL \\ 
        \hline
        \multirow{3}{*}{0\%} & Length & 3454.0 & 3812.0 & 3824.6 & \textbf{3855.4} (±45.4) \\ 
        ~ & Income & 27886.4 & 33110.3 & 31975.9 & \textbf{32912.5} (±1090.9) \\ 
        ~ & CpuAllo & 41.61\% & 45.0\% & 45.0\% & \textbf{45.2} (±0.5)\% \\ 
        \hline
        \multirow{3}{*}{30\%} & Length & 2786.9 & 2965.9 & 3003.8 & \textbf{3019.1} (±34.1) \\ 
        ~ & Income & 24133.7 & 27289.1 & \textbf{26935.9} & 26649.6 (±161.0) \\ 
        ~ & CpuAllo & 51.88\% & 53.5\% & 54.0\% & \textbf{54.1} (±0.2)\% \\ 
        \hline
        \multirow{3}{*}{40\%} & Length & 2439.8 & 2565.1 & \textbf{2604.0} & 2588.1 (±10.2) \\ 
        ~ & Income & 21270.0 & \textbf{22768.9} & 22448.6 & 22290.1 (±128.5) \\ 
        ~ & CpuAllo & 56.98\% & 57.9\% & \textbf{58.5\%} & 58.3 (±0.1)\% \\ 
        \hline
        \multirow{3}{*}{50\%} & Length & 1862.0 & 2078.4 & 2106.3 & \textbf{2181.8} (±9.1) \\ 
        ~ & Income & 14572.4 & 17456.7 & 17296.8 & \textbf{19108.4} (±75.3) \\ 
        ~ & CpuAllo & 61.55\% & 63.1\% & 63.5\% & \textbf{63.9} (±0.1)\% \\ 
        \hline
        \multirow{3}{*}{60\%} & Length & 1410.7 & 1502.9 & 1589.7 & \textbf{1602.1} (±5.7) \\ 
        ~ & Income & 10275.8 & 11009.4 & 12258.0 & \textbf{12417.8} (±24.5) \\ 
        ~ & CpuAllo & 67.18\% & 67.9\% & 68.5\% & \textbf{68.6} (±0.1)\% \\ 
        \hline
    \end{tabular}%
    }
\end{table}

\begin{table}[!ht]
    \centering
    \caption{Performance comparison of strategies across Expansion scenarios with 110 PMs. The experiments are conducted under the same conditions as those in Table 3.}
    \label{tb: performance-expansion}
    \resizebox{.8\linewidth}{!}{%
    \begin{tabular}{c|c|c|c|c|c}
        \hline
        $\rho_{ws}$ & Metric & First-Fit & Best-Fit & Internal-Scheduler & CVD-RL \\ 
        \hline
        \multirow{3}{*}{0\%} & Length & 3953.0 & 4366.6 & 4446.8 & \textbf{4475.5} (±40.0) \\ 
        ~ & Income & 34765.1 & 43570.6 & 44721.7 & \textbf{45303.2} (±618.5) \\ 
        ~ & CpuAllo & 59.36\% & 64.3\% & 64.6\% & \textbf{65.7} (±0.2)\% \\ 
        \hline
        \multirow{3}{*}{30\%} & Length & 3658.6 & 4083.1 & 4195.6 & \textbf{4229.4} (±22.9) \\ 
        ~ & Income & 33442.6 & 42080.8 & 44320.8 & \textbf{44557.6} (±329.0) \\ 
        ~ & CpuAllo & 63.42\% & 67.7\% & 68.2\% & \textbf{69.2} (±0.3)\% \\ 
        \hline
        \multirow{3}{*}{40\%} & Length & 3543.4 & 3965.3 & 4079.0 & \textbf{4085.9} (±29.2) \\ 
        ~ & Income & 32875.2 & 41593.3 & 41208.5 & \textbf{43343.5} (±531.7) \\ 
        ~ & CpuAllo & 65.17\% & 69.4\% & 69.8\% & \textbf{70.7} (±0.3)\% \\ 
        \hline
        \multirow{3}{*}{50\%} & Length & 3429.9 & 3818.6 & 3894.1 & \textbf{3910.8} (±16.9) \\ 
        ~ & Income & 32112.4 & 40612.0 & \textbf{42009.9} & 41888.3 (±420.4) \\ 
        ~ & CpuAllo & 67.09\% & 70.7\% & 71.3\% & \textbf{72.0} (±0.1)\% \\ 
        \hline
        \multirow{3}{*}{60\%} & Length & 3273.0 & 3599.3 & \textbf{3735.6} & 3707.6 (±28.1) \\ 
        ~ & Income & 30656.5 & 38239.0 & \textbf{40597.8} & 39641.6 (±375.6) \\ 
        ~ & CpuAllo & 68.61\% & 72.5\% & 73.1\% & \textbf{73.5} (±0.1)\% \\ 
        \hline
    \end{tabular}%
    }
\end{table}

\subsection{Traning Analysis}
\begin{figure}[htb!]
    \centering
    \begin{subfigure}[b]{0.45\textwidth}
        \centering
        \includegraphics[width=\textwidth]{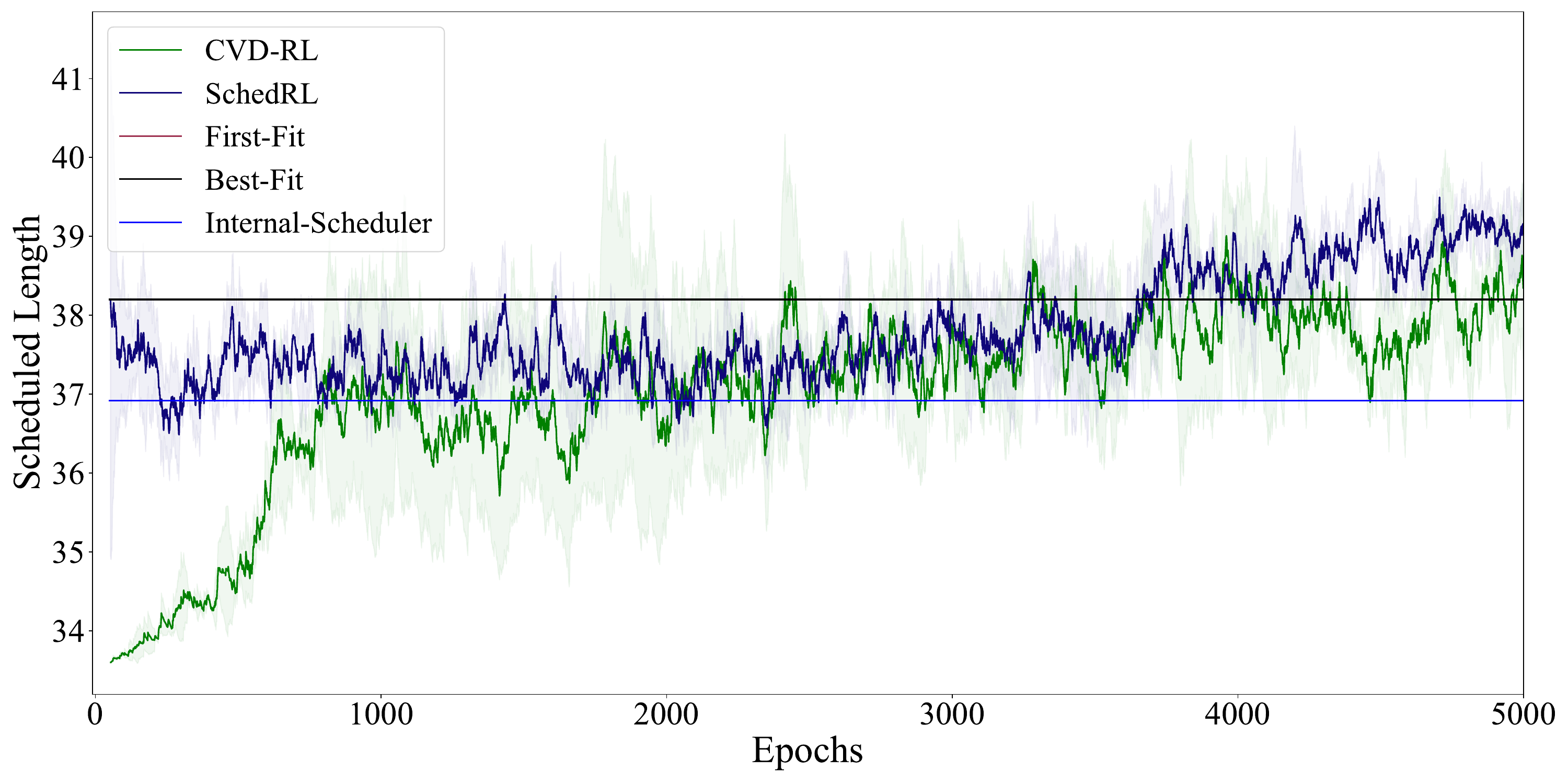}
        \caption{2 PMs.}
        \label{fig:training_curve_on_server2}
    \end{subfigure}
    \begin{subfigure}[b]{0.45\textwidth}
        \centering
        \includegraphics[width=\textwidth]{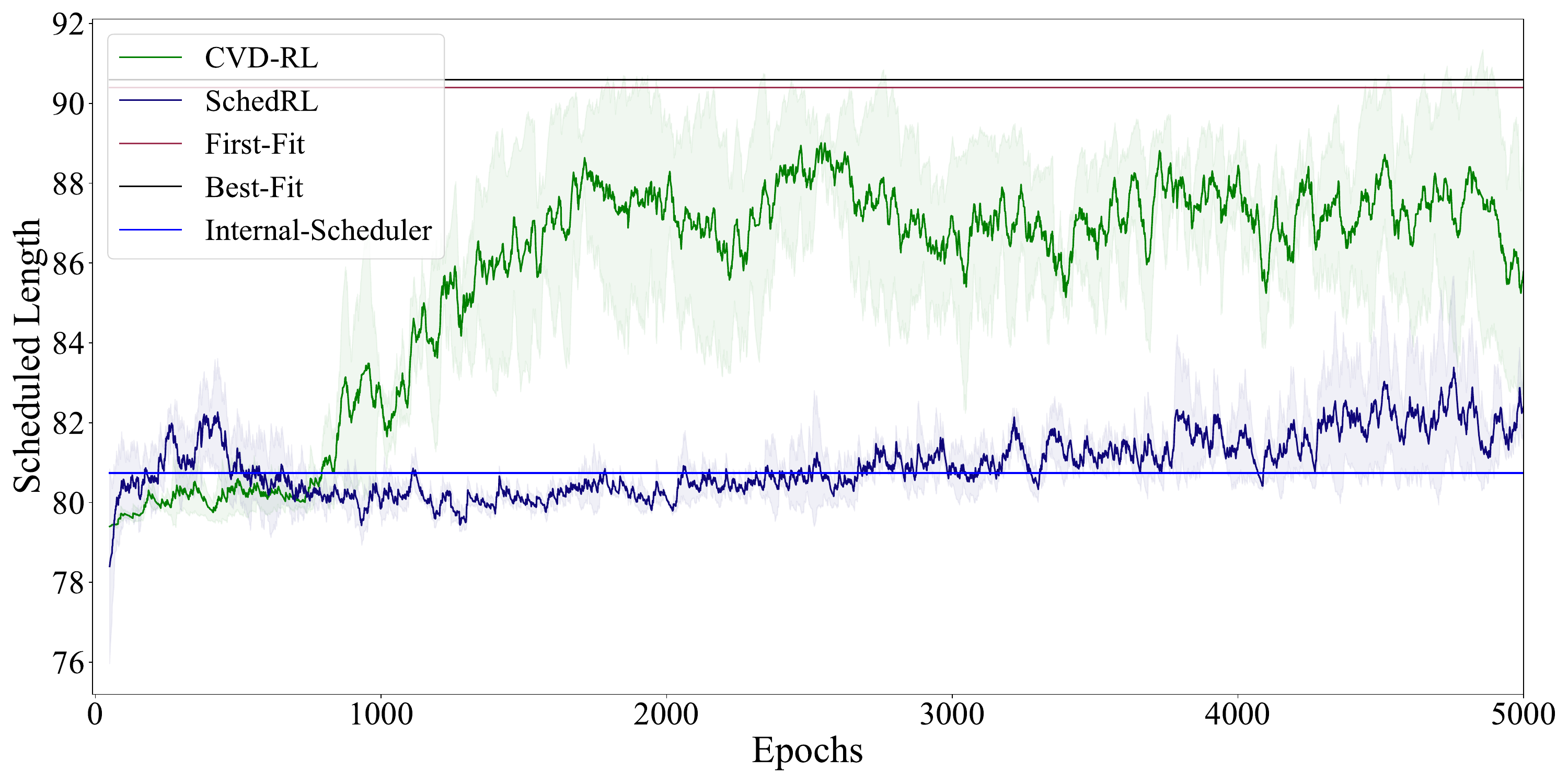}
        \caption{5 PMs.}
        \label{fig:training_curve_on_server5}
    \end{subfigure}
    \vskip\baselineskip
    \begin{subfigure}[b]{0.45\textwidth}
        \centering
        \includegraphics[width=\textwidth]{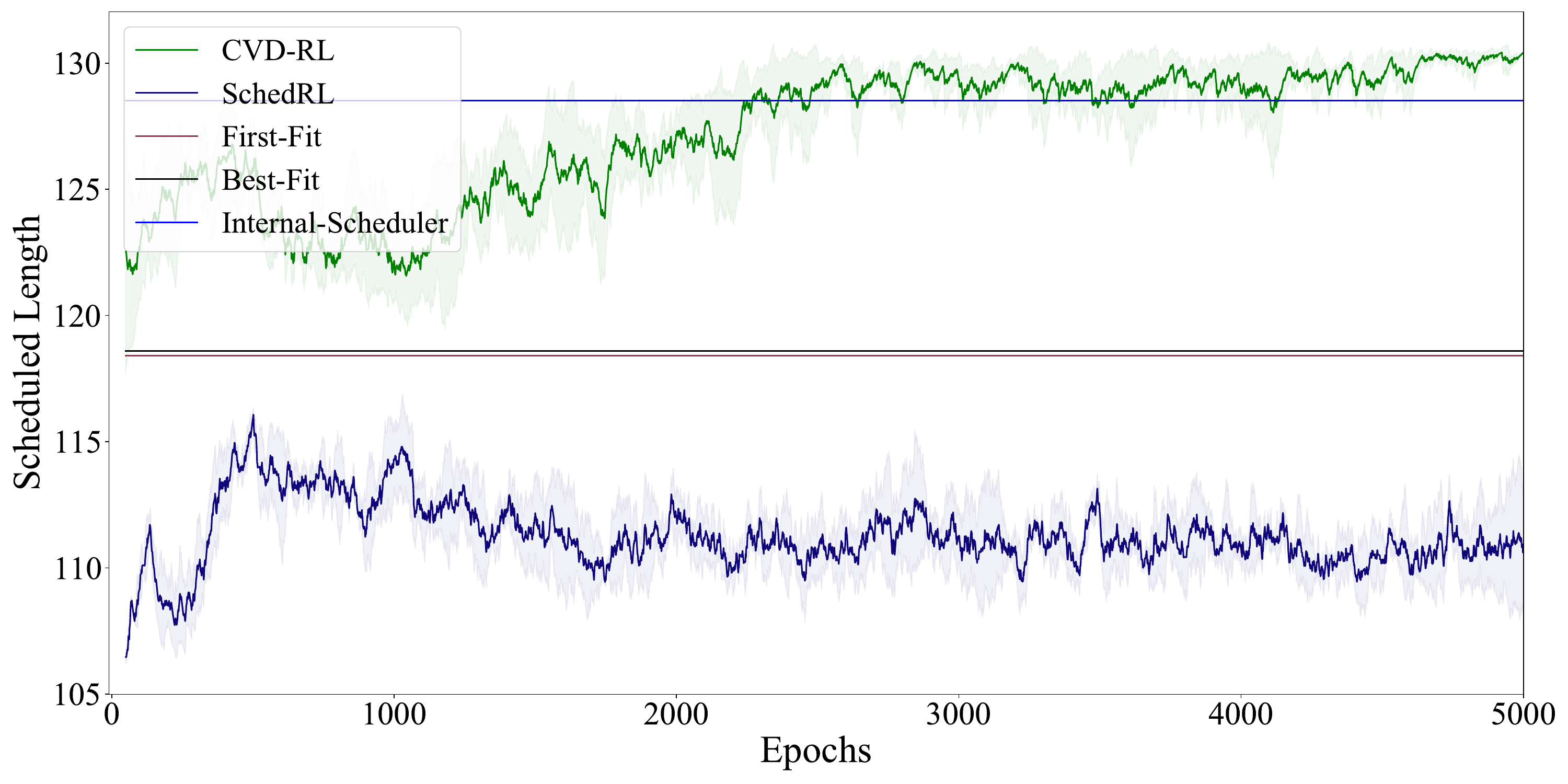}
        \caption{10 PMs.}
        \label{fig:training_curve_on_server10}
    \end{subfigure}
    \begin{subfigure}[b]{0.45\textwidth}
        \centering
        \includegraphics[width=\textwidth]{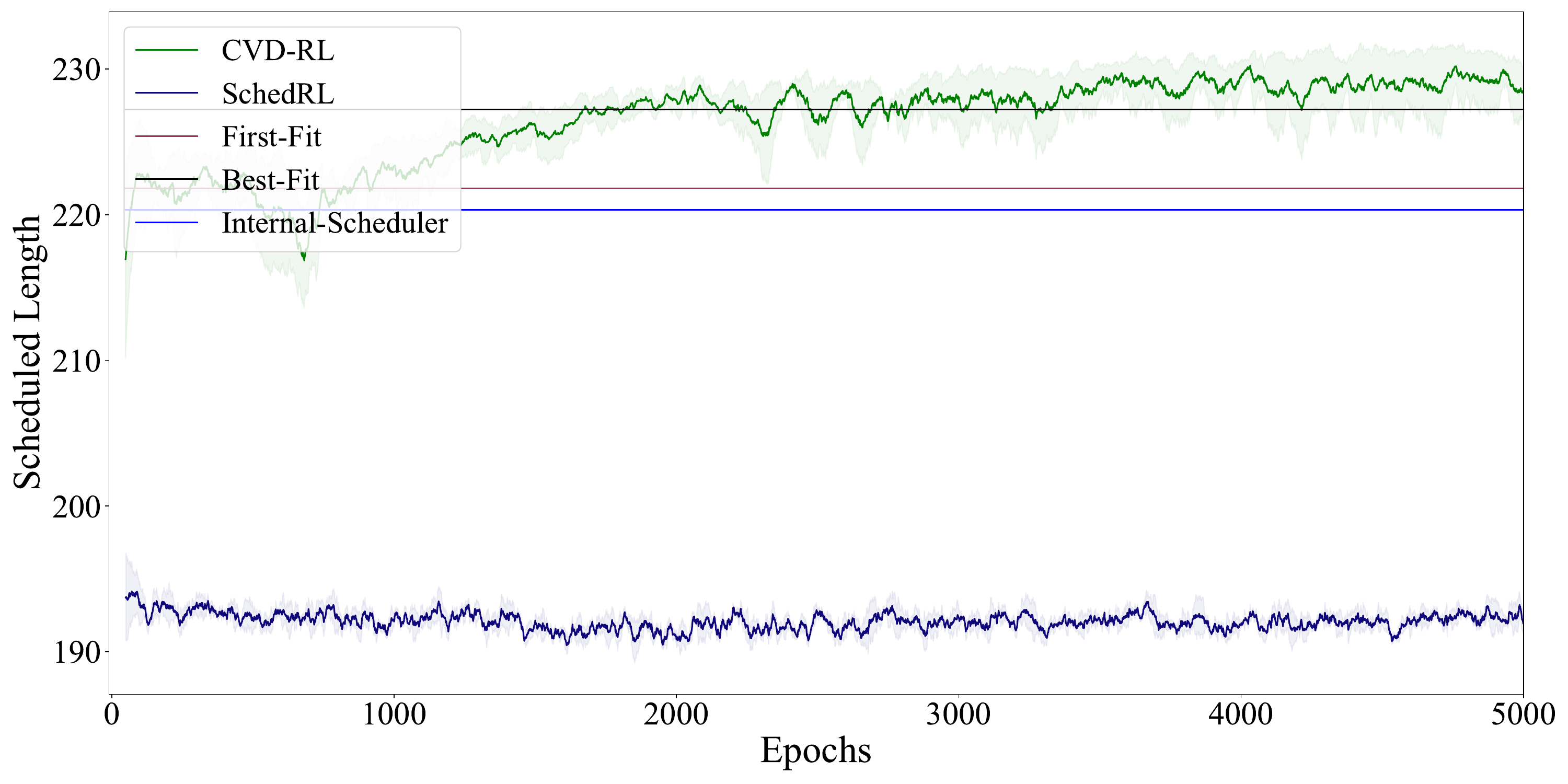}
        \caption{20 PMs.}
        \label{fig:training_curve_on_server20}
    \end{subfigure}
    \caption{Learning curves of CVD-RL and SchedRL for Non-Expansion with varying numbers of PMs.}
    \label{fig:training_curves_combined}
\end{figure}

We first train our CVD-RL in the Non-Expansion scenario with \(2\), \(5\), \(10\), and \(20\) PMs and \(0\%\) warm start ratio, using \(3000\) epochs and \(5\) different seeds. 
In each epoch, five parallel episodes are generated for the replay buffer. 
As shown in Fig.~\ref{fig:training_curves_combined}, our CVD-RL model achieves superior convergence and performance compared to SchedRL in environments with \(5\), \(10\), and \(20\) PMs, affirming its efficacy in different-sized clusters. 
The poor convergence of SchedRL in \(10\) and \(20\) PMs highlights the scalability issues in RL schedulers.

\begin{figure}[htb!]
    \centering
    \includegraphics[width=.8\linewidth]{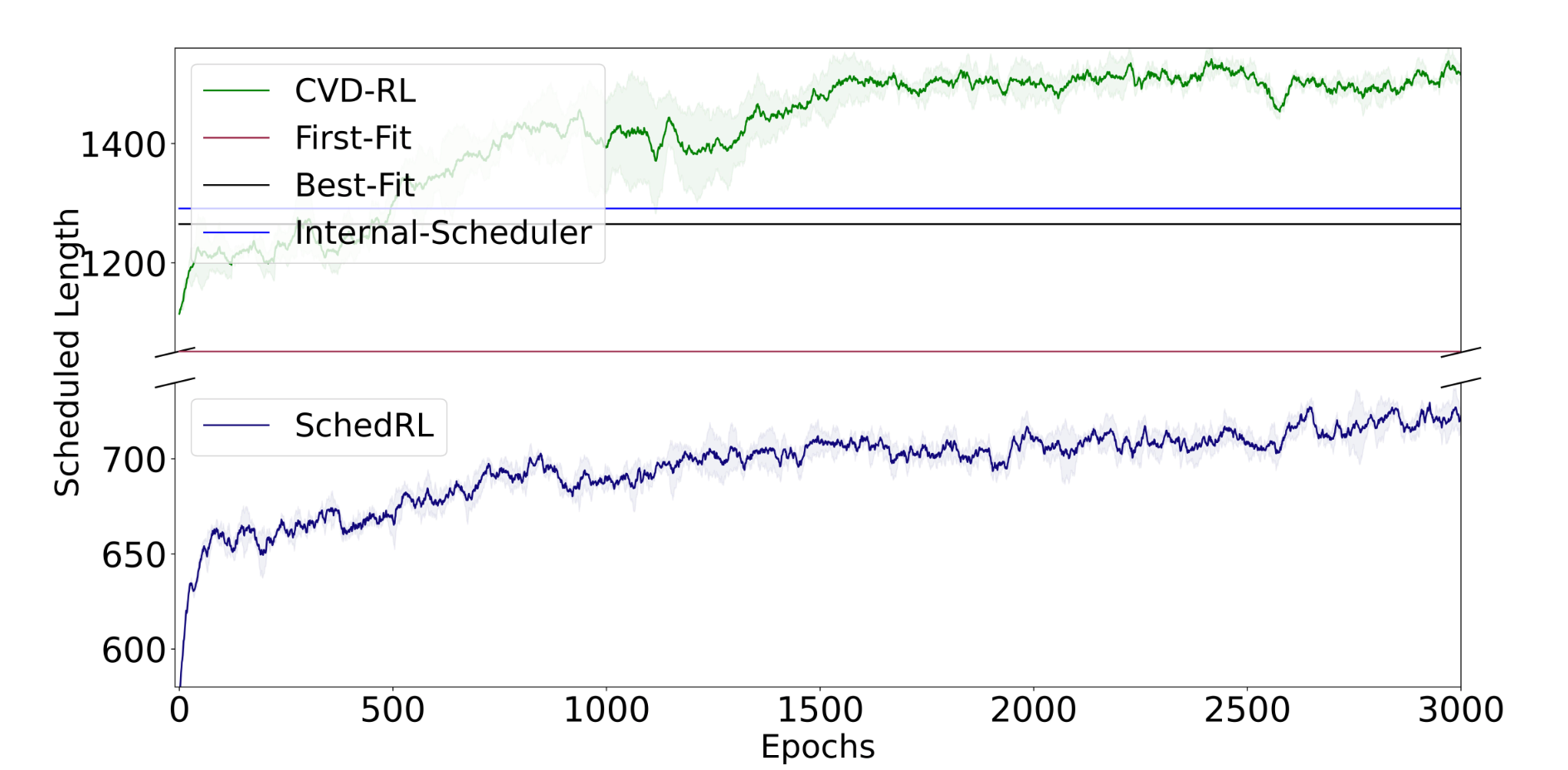}
    \caption{Learning curves for Scheduled Length in the Non-Expansion scenario with $50$ PMs and a $50$\% warm start ratio. The solid line represents the mean scheduled length, while the shaded area indicates the standard deviation.}
    \label{fig: baselines}
\end{figure}

Next, we consider a challenging Non-Expansion scenario with \(50\) PMs and a \(50\%\) warm start ratio. 
We train these methods for \(3000\) epochs with \(5\) different seeds. 
In each epoch, five parallel episodes are generated for the replay buffer. 
As depicted in Fig.~\ref{fig: baselines}, our approach surpasses baseline methods, demonstrating its superior effectiveness. 
Notably, SchedRL, despite its interactive learning paradigm, plateaus at a performance level below that of even the First-Fit method, underscoring the challenges in large-scale scheduling. 
Our model exhibits superior performance before reaching \(500\) epochs, a feat attributed to the implementation of a Top-$k$ filter, though it initially trails behind Best-Fit and Internal-Scheduler in terms of scheduled length. 
Beyond \(500\) epochs, it consistently outperforms all baseline models across performance metrics. 
We also tested these methods, and their performances are shown in Table~\ref{tb: performance-50}. 
CVD-RL outperforms all the baselines significantly in all three metrics for a \(50\%\) warm start ratio with \(50\) PMs. 
Note that even a \(1\%\) improvement can bring huge benefits due to the large-scale cloud \cite{hadary2020protean}.

\begin{figure}[htb!]
    \centering
    \begin{minipage}[b]{0.4\textwidth}
        \centering
        \includegraphics[width=\textwidth]{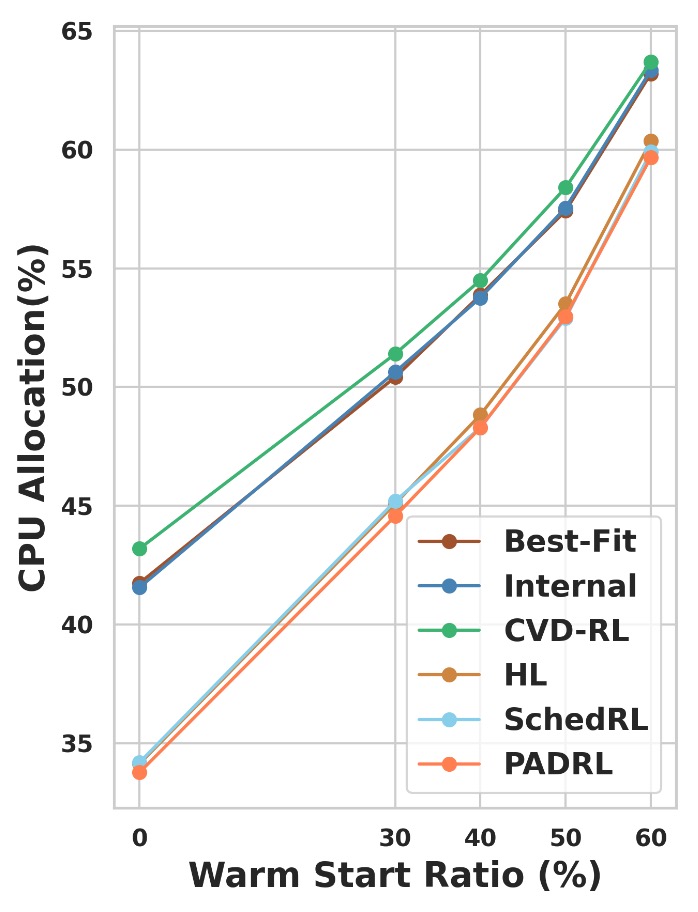}
        \label{fig: Non-Exp-cpuAllo}
    \end{minipage}
    \begin{minipage}[b]{0.4\textwidth}
        \centering
        \includegraphics[width=\textwidth]{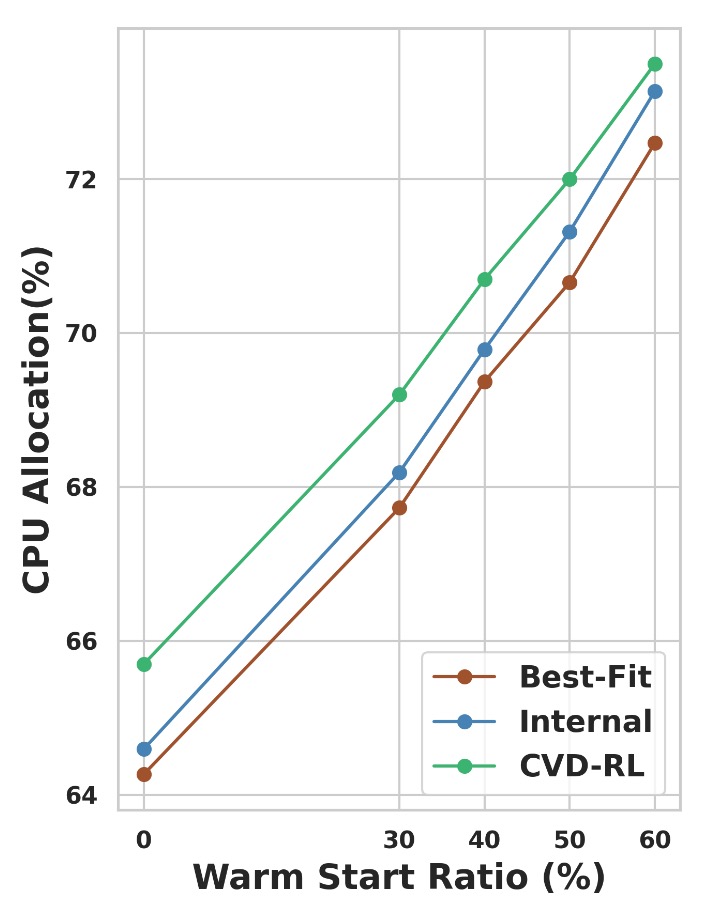}
        \label{fig: exp-cpuAllo}
    \end{minipage}
    \caption{Left: Non-Expansion; Right: Expansion. Average CPU Allocation. The left figure illustrates performance in the Non-Expansion scenario with $50$ PMs, whereas the right figure depicts CPU Allocation in the Expansion scenario.}
    \label{fig: cpuAllo}
\end{figure}

\subsection{Policy Generalization on Warm Starting}

We assess CVD-RL's generalization ability across different warm start ratios by applying the policy trained in the Non-Expansion scenario with \(50\) PMs and a \(50\%\) warm start ratio to other warm start ratios. 
As shown in Table~\ref{tb: performance-50}, CVD-RL outperforms all baselines across all considered warm start ratios and metrics. 
While learning-based methods typically perform worse than heuristic schedulers for \(50\) double-Numa PMs, CVD-RL, designed for scalability, surpasses heuristic schedulers. 
Specifically, CVD-RL exceeds First-Fit, Best-Fit, and Internal-Scheduler by at least \(6.4\%\), \(1.1\%\), and \(0.4\%\) in scheduled length, income, and average CPU utilization rate, respectively, at a \(60\%\) warm start ratio. 

These superior results across different warm start ratios demonstrate the model's applicability for various cluster statuses. Additionally, we visualize the CPU allocation improvement in Fig.~\ref{fig: cpuAllo}. 
This figure highlights CVD-RL's excellence in average CPU allocation, with more significant improvements observed at lower warm start ratios.

\subsection{Policy Generalization on Number of PMs}

Training a scheduling strategy for all numbers of PMs is intractable, and the cost of training in large-scale clusters is high. 
Therefore, investigating the applicability of our scheduling strategy to larger clusters is crucial for real-world implementation. 
The SchedRL, HL, and PADRL strategies struggle with generalization across different numbers of PMs, as their models are not naturally applicable to scenarios with a number of PMs different from those used in training. 
This section evaluates our model's scalability and generalization to clusters with \(100\) PMs. 
According to Table~\ref{tb: performance-100}, our strategy remains dominant in scenarios with a \(50\%\) warm start ratio and consistently outperforms baselines across most warm start ratios. 
For \(30\%\) and \(40\%\) warm start ratios, Best-Fit and Internal Scheduler achieve slightly higher results in specific metrics.

\begin{figure}[htb!]
    \centering
    \includegraphics[width=\linewidth]{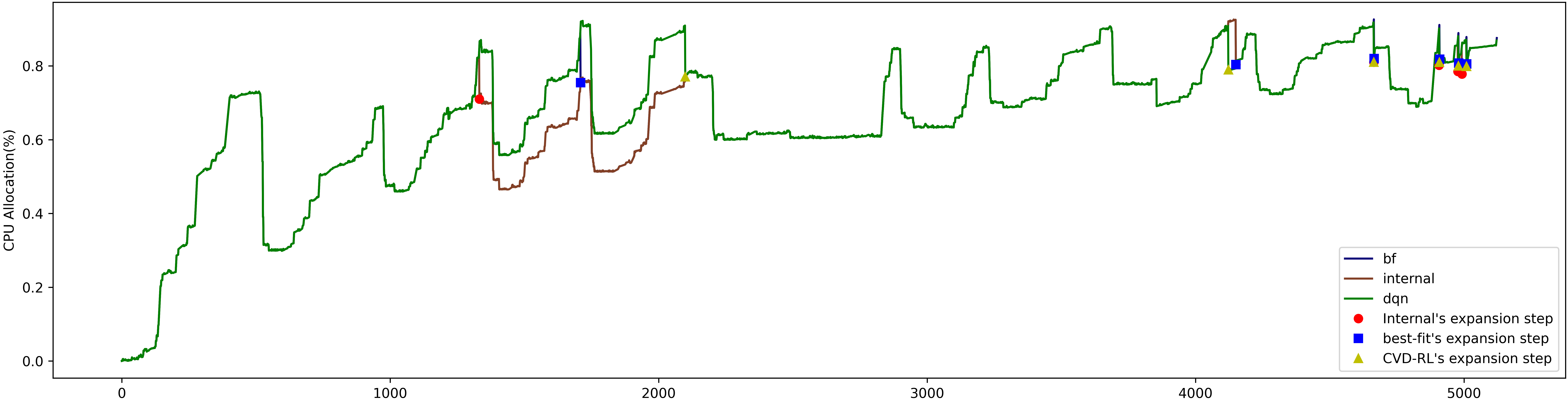}
    \caption{The CPU utilization in the Expansion scenario. Notably, CVD-RL demonstrates a deferred expansion step compared to Internal and Best-Fit. This could indicate a potential for better initial resource allocation or a different computational strategy employed by CVD-RL, resulting in its expansion step occurring at a later stage.}
    \label{fig:caseForAblations}
\end{figure}

We also include the expansion scenario in testing, simulating the continuous growth in cluster capacities faced by cloud service providers. 
As shown in Table~\ref{tb: performance-expansion}, our model effectively addresses the dynamic challenges posed by expanding infrastructure. 
Notably, other learning-based schedulers are inapplicable in this scenario due to their lack of adaptability to changing PM counts. 
Table~\ref{tb: performance-expansion} and the right figure in Fig.~\ref{fig: cpuAllo} collectively demonstrate our model's consistent superiority in average CPU allocation within the Expansion scenario, achieving at least a \(0.7\%\) improvement over alternatives. 
However, the Internal-Scheduler manages a longer scheduled length in scenarios with a \(60\%\) warm start ratio than our CVD-RL model. 
This discrepancy, especially in scheduled length and CPU allocation rate, may be attributed to differences in expansion frequencies. 
To this end, we analyze the CPU allocation dynamics across different methods, further visualized in Fig.~\ref{fig:caseForAblations}, providing insights into the operational efficiencies and scheduling decisions underpinning these observations.

\subsection{Ablation Studies}

This section explores the impact of various components within CVD-RL. 

\subsubsection{Operator-Specific Impact Analysis}

\begin{figure}[htb!]
    \centering
    \includegraphics[width=0.8\textwidth]{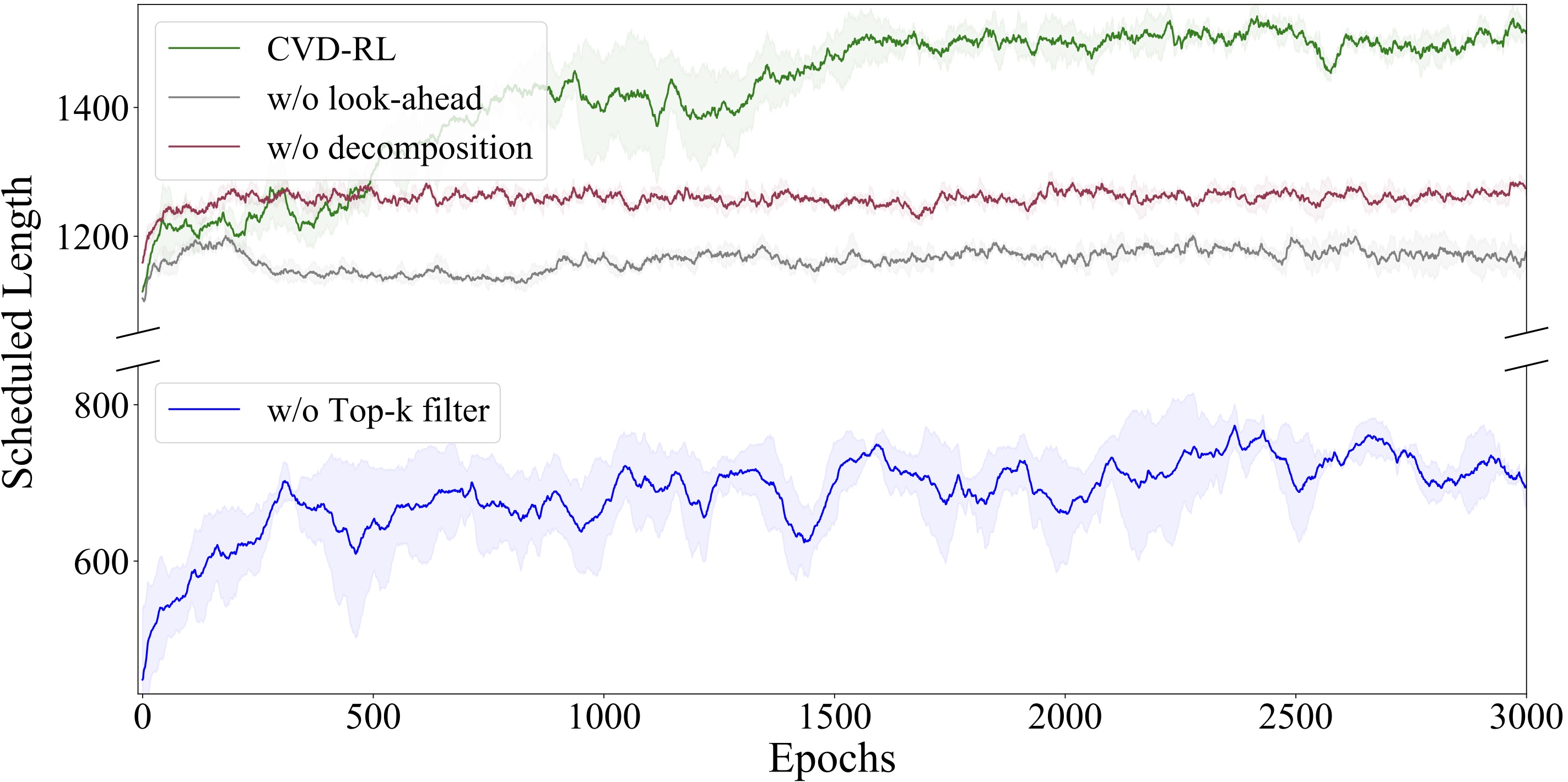}
    \caption{Impact of operator removal on the scheduled length in the Non-Expansion scenario with $50$ PMs and a $50$\% warm start ratio. The solid line indicates the mean scheduled length, while the shaded area represents the standard deviation.}
    \label{fig: ablations}
\end{figure}

To assess the importance of individual operators, we conducted experiments by systematically omitting each one during the training process. 
Fig.~\ref{fig: ablations} depicts the negative impact on performance resulting from the removal of any operator, highlighting their collective significance in our model.

The Top-$k$ Filter Operator is identified as critically essential. 
Removing this operator led to a dramatic decrease in CVD-RL's performance, with the scheduled length dropping to below $500$, in stark contrast to its performance when fully equipped. 
This significant decline underscores the importance of efficient exploration within a large action space and the necessity of dynamic action space reduction techniques in RL.
Similarly, excluding the decomposition or look-ahead operators resulted in a noticeable decline in performance, with scheduled lengths falling below $1300$, as opposed to exceeding $1400$ with the complete CVD-RL model. This demonstrates the vital role these operators play in facilitating optimal scheduling outcomes.

\subsubsection{Exploring the Top-k Filter Operator Design}

\begin{figure}[htb!]
    \centering
    \includegraphics[width=0.8\textwidth]{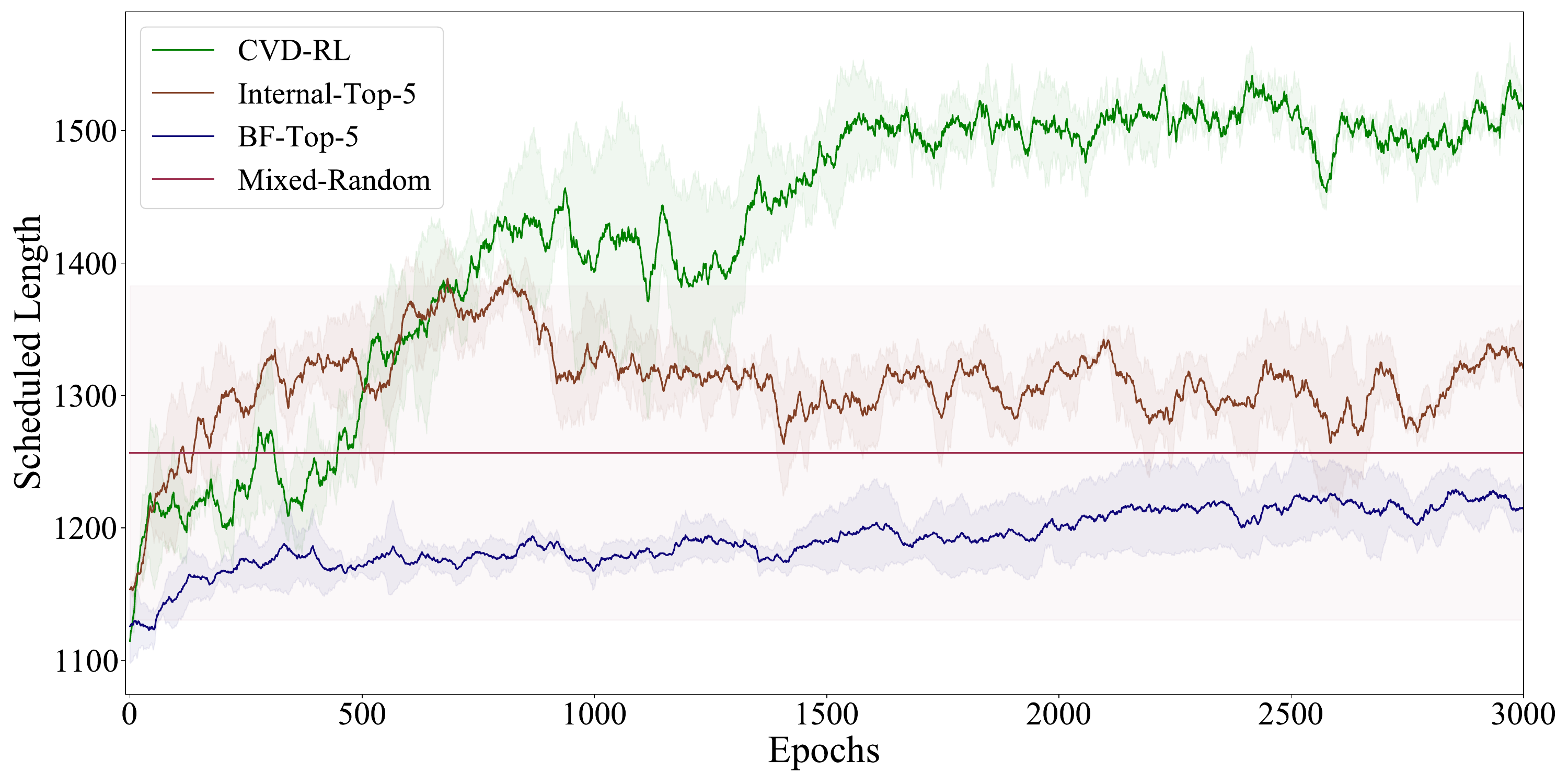}
    \caption{Performance comparison of CVD-RL using different Top-5 Filters in the Non-Expansion scenario with $50$ PMs and a $50$\% warm start ratio.}
    \label{fig: comparative_on_filter}
\end{figure}

Given the pivotal role of the Top-$k$ Filter Operator in our approach, we investigate the effects of its various implementations on the scheduler's performance. 
We compared filters modeled after the Best-Fit and our Internal-Scheduler algorithms, generating five candidate actions each, denoted as BF-Top-5 and Internal-Top-5, respectively. 
Fig.~\ref{fig: comparative_on_filter} shows that all filters improve performance over the baseline, with our CVD-RL model achieving the most significant enhancement. 
This suggests that leveraging a diverse set of strategies for generating candidate actions allows our model to excel.

\begin{figure}[htb!]
    \centering
    \includegraphics[width=0.8\textwidth]{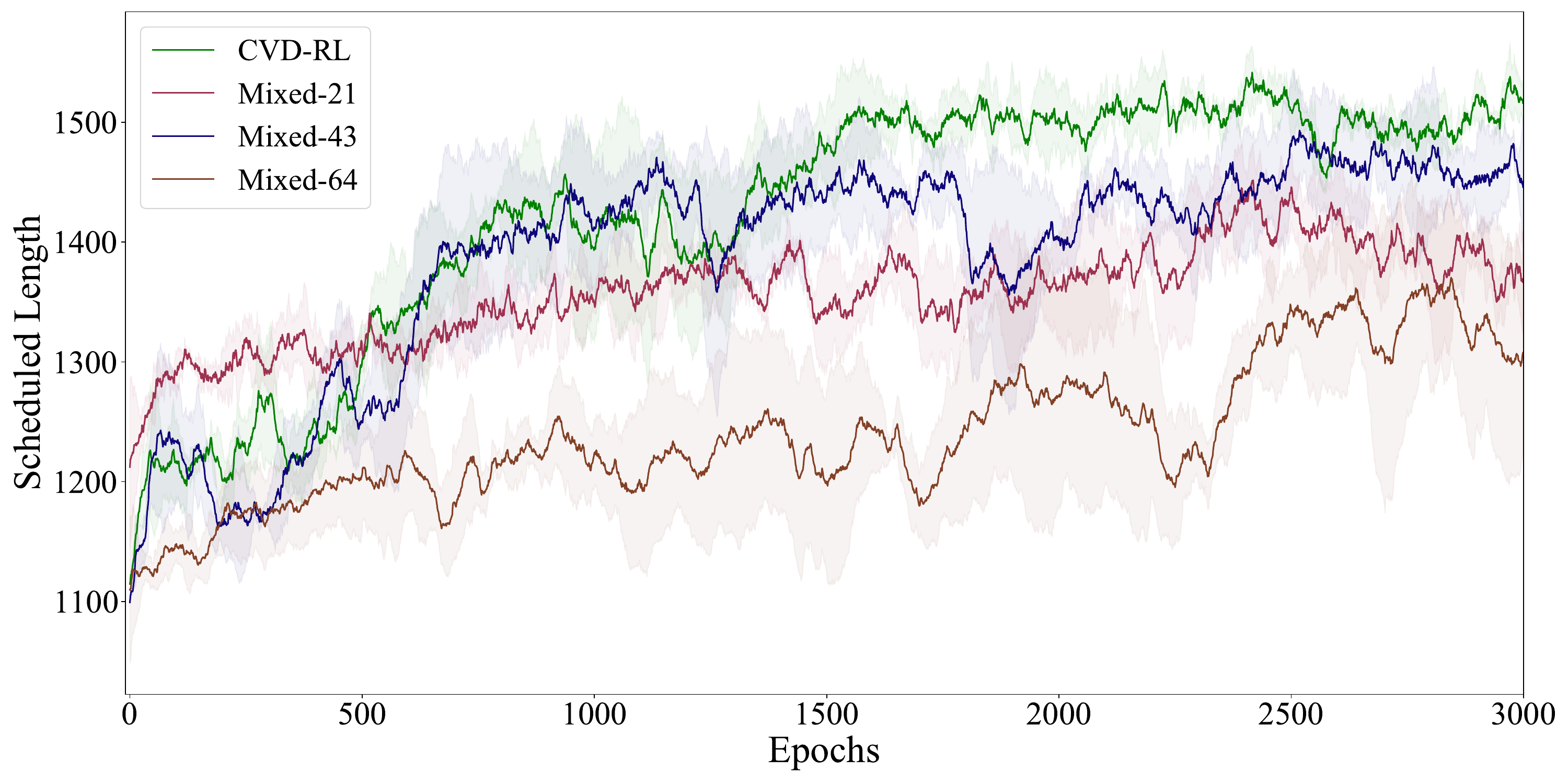}
    \caption{Learning curves for CVD-RL with varying $k$ values for Top-$k$ Filters in the Non-Expansion scenario with $50$ PMs and a $50$\% warm start ratio.}
    \label{fig: comparative_on_K}
\end{figure}

Furthermore, the choice of $k$ in the Top-$k$ Filter plays a critical role. 
Lower values of $k$ streamline the action space but may hinder thorough exploration. 
Achieving optimal performance necessitates striking a balance between limiting and exploring the action space. 
We tested the CVD-RL model with $k$ values set to $3$, $5$, $7$, and $10$, resulting in the Mixed-21 (Top-3 filter), Mixed-43, and Mixed-64 configurations, respectively. 
These configurations were evaluated over $3000$ epochs with $50$ PMs, with their learning curves presented in Fig.~\ref{fig: comparative_on_K}. 
The Mixed-21 model, leveraging the strengths of both the Internal-Schedule and Best-Fit, initially exhibited superior performance due to its more focused action space. 
However, an overly constrained action space led to limited exploration and potential entrapment in local optima. 
On the other end, the Mixed-64 model showed less effective convergence and final performance, indicating that an excessively broad search space can impede efficient learning. 
This highlights the crucial balance provided by the Top-$k$ filter in RL for VMS, with our Mixed-32 approach (equivalent to the original model) demonstrating the optimal balance between performance and convergence.

\section{Related Works}

This section reviews existing literature on VMS and RL applications for cloud resource scheduling.

\subsection{Virtual Machine Scheduling}
\label{sec:related-work-vms}
VMS is a critical aspect of cloud resource management. 
The primary objective of VMS is to enhance resource utilization and maximize the number of processed requests with a given set of PMs. 
This task is recognized as an NP-hard problem, necessitating alternative solution strategies~\cite{salot2013survey}.

Researchers have proposed various heuristic methods for theoretical analysis, focusing on competitive ratios. 
Stolyar et al.\cite{stolyar2013infinite} first proved that the greedy algorithm is asymptotically optimal under certain conditions on VM request distribution. 
Li et al.\cite{li2015dynamic} formulated the VMS problem as a dynamic bin packing problem. 
Dosa et al.\cite{dosa2013first} analyzed the competitive ratio of First-Fit, and Dosa et al.\cite{dosa2014optimal} analyzed the competitive ratio for Best-Fit in the dynamic bin packing problem. 
However, these approaches often rely on stringent assumptions, such as the Poisson distribution of VM requests in~\cite{stolyar2013infinite}.

In practical scheduling, heuristic methods are prevalent in the industry. 
Cloud computing companies typically use variants of First-Fit and Best-Fit to construct their internal schedulers~\cite{hadary2020protean, sheng2022learning}, incorporating rules proposed by domain experts. 
However, these heuristic methods cannot guarantee optimal solutions.

Another approach to developing practical scheduling policies involves RL. 
Researchers formulate the VMS problem as a MDP. 
Sheng et al.\cite{sheng2022vmagent} proposed an RL-friendly simulator for training RL methods to learn scheduling policies and subsequently developed a RL method named SchedRL~\cite{sheng2022learning} for small-scale environments. 
However, SchedRL faces significant challenges in large-scale environments due to the exponentially increasing state-action space, and its performance is often inferior to Best-Fit. 
Lastly, Sinclair et al.\cite{sinclair2023hindsight} proposed hindsight learning, which allows schedulers to imitate hindsight heuristic schedulers. 
Although not an RL method, its performance is constrained by the hindsight scheduler.

In contrast to these works, we propose a scalable RL method for VMS that leverages both trial-and-error learning and scalability features.

\subsection{RL for Cloud Resource Scheduling}
RL has emerged as a potent tool in addressing various cloud resource scheduling problems. 
Its applications range from VM rescheduling to job scheduling and VMS.

Job scheduling studies when and which jobs should be allocated to VMs~\cite{mao2016resourcemanagement}. 
Each job is characterized by its runtime, size, and priority. 
\cite{mao2016resourcemanagement} first proposed DeepRM, a DRL-based approach for job scheduling. 
DeepRM translates the packing tasks with multiple resource demands into a learning problem, demonstrating comparable performance to SOTA heuristics. 
\cite{mao2019learning} further introduced Decima. Decima integrates a graph neural network to extract job DAGs and cluster status as embedding vectors, feeding these to a policy gradient network for decision-making. 
They scale RL for job scheduling in the cloud to $25$ VMs. 
\cite{li2024batch} extended the scenario to batch job scheduling in cloud computing using distributional RL. 
They designed a system model that includes complex batch jobs arriving over time, dynamically changing multidimensional cluster resources, and a scheduler that maintains load balancing. 
However, it is evaluated within $9$ VMs. 
\cite{fan2022dras} proposed DRAS to improve scalability. 
They adopted two-level neural networks to select the job handling scheme instead of selecting each job's location.

VM rescheduling involves migrating allocated VMs from their current PMs to other PMs to optimize resource usage while satisfying service level agreements~\cite{ding2022vmr2l}. 
\cite{raven} first employs DRL to choose destination PMs for VMs during migration. 
However, their methods are difficult to converge when the PM number increases to $50$.
\cite{hummaida2022scalable} propose a decentralized RL approach to enhance the scalability of VM migration in large-scale cloud environments. 
It groups the PMs based on CPU utilization and significantly reduces the action space. 
\cite{zeng2022adaptive} propose the Prediction Aware DRL-based VM placement method (PADRL) framework which  use LSTM to provide DRL-based model reasonable environment state.
\cite{ding2022vmr2l} employ a two-stage framework and propose a novel feature extraction method to improve scalability. Recently, \cite{fukushima2024application} propose to decompose the VM migration control problem into determining VM distribution across edge servers and determining the exact locations of VMs. They adopt DDPG~\cite{lillicrap2015continuous} to determine VM distribution while using a heuristic algorithm to determine the exact locations.

VMS involves selecting PMs to handle online VM requests. 
Compared to VM rescheduling and job scheduling, it has less information about the VM and PM features. 
This makes it challenging to improve scalability based on these features. 
As introduced in \ref{sec:related-work-vms}, although RL for VMS methods shows promising results in small-scale environments, scaling RL to large-scale VMS problems is still under-explored, and our work makes an attempt in this direction.

Our work also relates to multi-agent reinforcement learning (MARL). 
MARL studies how RL can be applied to problems involving multiple agents~\cite{coma,sunehag2017value,mappo,leibo2021scalable}. 
The exponentially increasing state-action space is also common in MARL. 
Due to the problem's natural decomposition into individual agent problems, researchers can design effective methods to address the scalability issue. 
However, the decision-making in the VMS problem is strongly coupled. 
Unlike MARL, our work decomposes the representation of the value function within the single-agent RL.

\section{Conclusion}
This paper introduced the CVD-RL framework, a novel approach to VMS in large-scale cloud computing environments. 
It proposes three operators to increase the scalability of RL in VMS. 
Empirical evaluations demonstrate CVD-RL's superior performance in CPU allocation efficiency and its ability to handle clusters of up to $50$ PMs. 
Moreover, the obtained strategy shows consistent performance when scaled to clusters with a larger number of PMs, different initialization, and even expansion scenarios. T
his offers a promising solution to the challenges faced in contemporary cloud computing resource management.
Regarding future directions, exploring methods to enhance the scalability of RL in VMS presents an intriguing avenue of research. 
Specifically, the application of hierarchical RL, curriculum learning, and other advanced techniques holds potential in this direction.
Rescheduling is also a promising direction.
In this setting, VMs is allowed to migrate, which means reassigning them to different PMs.
We believe these lines of research complement each other and may be combined in future efforts.

\bibliography{aaai25}

\clearpage
\newpage

\appendix

\section{Proofs}

\subsection{Proof of the Decomposition Operator}
\label{proof: decomp}

This proof demonstrates that after applying the decomposition operator, the Best-Fit policy, denoted as \(\pi^{\mathrm{bf}}\), remains within the policy space. The Best-Fit policy can be defined as follows:

\[
\pi^{\mathrm{bf}}(\boldsymbol{a} | \boldsymbol{s}) := \arg \max_{\boldsymbol{a}\in \mathbb{A}} h(\boldsymbol{s}, \boldsymbol{a}),
\] where \(h\) is the Best-Fit's score function, which can be decomposed as 

\[
h(\boldsymbol{s}, \boldsymbol{a}) = \sum_i h_i(s_i^c, \boldsymbol{s}^v, \bar{a}_i),
\] where \(h_i\) is the PM-wise Best-Fit score function. Then we have \(i^* = \arg \max_i \max_{\bar{a}_i \in \bar{\mathbb{A}}} h_i(s_i^c, \boldsymbol{s}^v, \bar{a}_i)\) and 

\[
\bar{a}_{i}^* = \left\{ \begin{array}{ll}
    [0,0], & \text{if } i \neq i^*,\\
    \arg \max\limits_{\bar{a}_{i} \in \bar{\mathbb{A}}_i} h_i\big( s_i^c(t), \boldsymbol{s}^v, \bar{a}_i \big), & \text{otherwise},
    \end{array}\right.
\]

The policy selects the action that maximizes the heuristic function \(h\) for the PM with the highest \(h_i\), effectively implementing the Best-Fit strategy by prioritizing the allocation on the PM that best fits the requirements of the incoming request.

The condition that ensures the Best-Fit policy is captured within the policy space is:

\[
\bar{Q}_i \big( s_i^c, \boldsymbol{s}^v, \bar{a}_i; \bar{\theta} \big) - \bar{Q}_i \big( s_i^c, \boldsymbol{s}^v, \hat{a}_i; \bar{\theta} \big) = h_i\big( s_i^c, \boldsymbol{s}^v, \bar{a}_i \big),
\] for all \(s_i^c\), \(\boldsymbol{s}^v\), and \(\bar{a}_i \in \bar{\mathbb{A}}_i \setminus \hat{a}_i\). 
This equation signifies that the Q-value difference for selecting action \(\bar{a}_i\) over the default action \(\hat{a}_i\) equates to the heuristic value \(h_i\) of that action, thus aligning the Q-learning objective with the Best-Fit heuristic. 
When this condition is satisfied across all PMs and potential actions, the Best-Fit policy is achieved within the decomposed policy space, confirming that the decomposition operator preserves the feasibility of implementing Best-Fit as a viable policy.

\end{document}